\pdfoutput=1
\documentclass[11pt,letterpaper]{extarticle}

\usepackage{xcolor}

\usepackage[libertine]{hug_arxiv}
\runninghead{HuG: Learning Humanoid Whole-Body Affordance-Aware Grasping with Active Perception}
\venue{Conference on Robot Learning (CoRL) 2026, Austin, TX, USA}
\hypersetup{pdftitle={HuG: Learning Humanoid Whole-Body Affordance-Aware Grasping with Active Perception},pdfauthor={Xuanbin Peng, Lai Wei, Yuze Zhai, Qingyuan Yu, Keyi Wang, Jianglong Ye, Xiaolong Wang}}

\usepackage{amsmath}
\usepackage{amssymb}
\usepackage{bbm}
\usepackage{graphicx}
\usepackage{multirow}
\usepackage{float}
\usepackage{pifont}

\newcommand{\bT}{\mathbf{T}}
\newcommand{\ba}{\mathbf{a}}
\newcommand{\bh}{\mathbf{h}}
\newcommand{\bp}{\mathbf{p}}
\newcommand{\bq}{\mathbf{q}}
\newcommand{\br}{\mathbf{r}}
\newcommand{\bs}{\mathbf{s}}
\newcommand{\bv}{\mathbf{v}}
\newcommand{\bz}{\mathbf{z}}
\newcommand{\bmu}{\boldsymbol{\mu}}
\newcommand{\bsigma}{\boldsymbol{\sigma}}
\newcommand{\bepsilon}{\boldsymbol{\epsilon}}
\DeclareMathOperator{\SDF}{SDF}

\newcommand{\secv}{}
\newcommand{\ssecv}{}

\title{HuG: Learning Humanoid Whole-Body Affordance-Aware Grasping with Active Perception}

\author{
  \normalfont Xuanbin Peng$^{*\,\S}$, Lai Wei$^{*}$, Yuze Zhai$^{*}$, Qingyuan Yu$^{*}$, Keyi Wang\\
  Jianglong Ye$^{\dag}$, Xiaolong Wang$^{\dag}$\\[2pt]
  UC San Diego\\
  \href{https://xuanbinpeng.com/hug}{\texttt{xuanbinpeng.com/hug}}
}

\begin{document}
\maketitle
{\let\thefootnote\relax\footnotetext{$^*$Equal contribution.\quad $^\S$Project lead.\quad $^\dag$Equal advising.}}

\begin{figure}[!h]
    \centering
    \includegraphics[width=\linewidth]{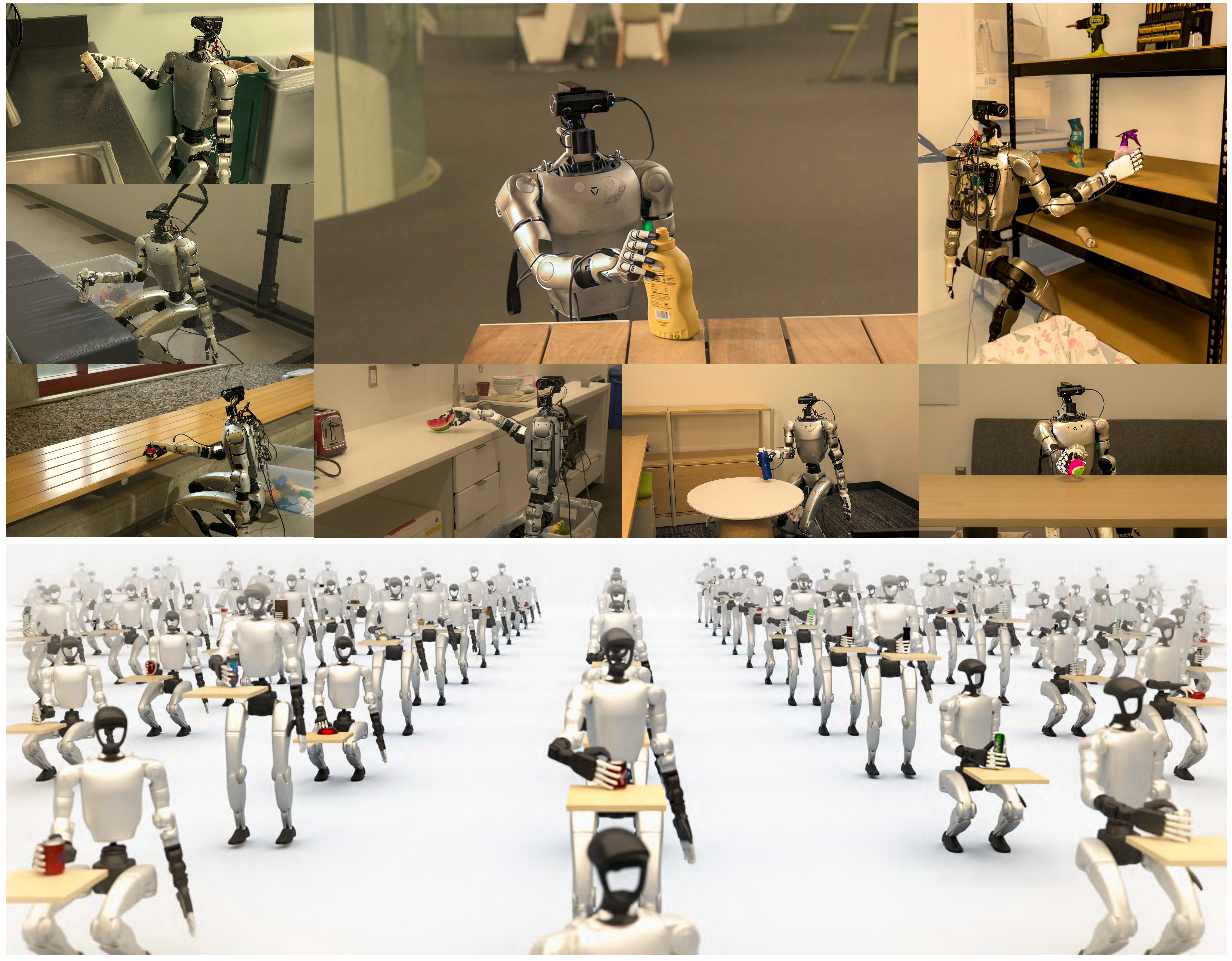}
    \caption{Humanoid whole-body affordance-aware grasping.
    Top: our system runs zero-shot on a Unitree G1, coupling active perception with whole-body control to discover and grasp objects across diverse everyday scenes, from deep squats at low tables to full extension at high shelves.
    Bottom: the policy is trained entirely on large-scale synthetic data generated and filtered with kinodynamic feasibility.}
    \label{fig:teaser}
\end{figure}

\secv
\begin{abstract}
Acquiring objects in human environments is hard for humanoids: 
targets are partially occluded under egocentric viewpoints, and reaching them demands coordinated whole-body motion 
that keeps the target visible and reachable while the base stays stable. We present HuG, a framework for whole-body affordance-aware 
grasping with active perception that couples perception and action through a Perceive--Align--Grasp pipeline: belief-driven active 
perception resolves occlusion by treating viewpoint as an explicit decision variable; a point-cloud--conditioned affordance policy 
proposes whole-body grasps; and an affordance-aware alignment stage repositions the body to the most executable configuration. 
A key insight is that, on a floating base, \emph{kinematic validity is not enough}: 
we curate training data with controller-in-the-loop dynamics simulation, rejecting 73\% of kinematically valid grasps
 that would destabilize the deployment controller, so the policy learns kinodynamically feasible affordances.
 Trained entirely in simulation with zero real demonstrations, HuG attains 87.5\% real-world success on a multi-level shelf task, 
 generalizes to novel objects, and holds ${\ge}$73\% at every tested height. 
 Beyond generalizing within its training distribution, the policy extrapolates spatially: at scale in simulation 
 it succeeds at table heights well below any seen during training, indicating that kinodynamically filtered synthetic data grounds affordances 
 in object geometry and whole-body feasibility rather than in memorized configurations. 
\end{abstract}

\secv
\section{Introduction}
\secv
\label{sec:introduction}

Humanoid robots promise to bring general-purpose object acquisition into human environments.
 Yet in realistic homes and labs, targets are rarely presented in a clean tabletop configuration:
  they are frequently \emph{partially occluded} under egocentric viewpoints and spread across diverse spatial configurations, 
  and reaching them demands \emph{coordinated whole-body motion}---turning the head, bending the torso, squatting---to maintain visibility, 
  reachability, and stability at once. 
  Consider retrieving an item from a cluttered multi-level shelf: the target may be visible only from a specific angle, 
  sit at a height requiring a deep squat or a full extension, and afford a stable grasp only from particular approach directions. 
  Such tasks demand tightly coupling perception with whole-body action: not as independent modules, 
  but as stages in which viewpoint, posture, and grasp decisions jointly shape one another.

Three threads of prior work have each made striking progress on this coupling, yet each leaves a critical gap.
\emph{(i)~Imitation from real demonstrations} scales humanoid manipulation through teleoperation~\citep{ben2025homiehumanoidlocomanipulationisomorphic,ze2025twist2scalableportableholistic,he2024learning,li2025cloneclosedloopwholebodyhumanoid}, with ACT~\citep{zhao2023learning}-style architectures and diffusion policies such as ViA~\citep{xiong2025via} supporting long-horizon tasks; but these policies are bounded by their demonstration coverage, generalizing poorly to objects and layouts outside it; worse, on a floating base every action error also shifts the body, so errors compound over long-horizon rollouts into end-effector drift that centimeter-level grasping cannot absorb.
\emph{(ii)~Large-scale synthetic learning} enables zero-shot sim-to-real via end-to-end policies~\citep{he2025viral,xue2025openingsimtorealdoorhumanoid,lin2025sim,wei2026psi0openfoundationmodel} and modular grasp synthesis~\citep{fang2020graspnet,ye2025dex1b,wang2023dexgraspnet,zhang2024dexgraspnet20learninggenerative,xu2023unidexgrasp,deng2025graspvlagraspingfoundationmodel,dong2026herolearninghumanoidendeffector}; but end-to-end humanoid policies reason about dynamics only \emph{implicitly}, while grasp planners are built for \emph{fixed-base} tabletops, so a \emph{kinematically valid} grasp can be \emph{dynamically infeasible} once a floating-base humanoid must execute it under whole-body control. Neither line explicitly enforces \emph{kinodynamic whole-body feasibility} during data generation.
\emph{(iii)~Active perception} regulates viewpoints, from classical active perception and active SLAM~\citep{bajcsy1988active,placed2023surveyactivesimultaneouslocalization} to learned perceptual strategies~\citep{xiong2025via,jangir2022look,hu2025realworldreinforcementlearningactive,luo2025emergent}; but on humanoids, viewpoint regulation either emerges implicitly from demonstrations, bounded by their coverage~\citep{xiong2025via}, or is confined to the head with the body left passive~\citep{qi2025coordinated}. Yet a body-attached camera binds viewpoint to whole-body posture~\citep{kim2024armoregocentricperceptionhumanoid,liu2024visual,yin2025womapworldmodel}, so resolving occlusion can require lowering the torso or leaning, not merely looking.
Humans, by contrast, couple perception and whole-body action instinctively: they lean past occlusions, squat to low shelves, and turn to approach from stable angles. Current humanoid grasping systems lack this coupling.

We present \textbf{HuG}, a hierarchical framework for whole-body affordance-aware grasping with active perception (Fig.~\ref{fig:teaser}) that realizes this coupling through a Perceive--Align--Grasp pipeline:
\textbf{(1)~belief-driven active perception}, a Bayes-style belief filter over an egocentric viewpoint grid that resolves occlusion by systematic search, treating viewpoint as an \emph{explicit} decision variable rather than emergent behavior;
\textbf{(2)~kinodynamically filtered affordance learning}, where an \emph{affordance} is a kinodynamically feasible whole-body grasp configuration for the observed object: a point-cloud-conditioned CVAE is trained on 600k synthetic trajectories that each pass \emph{controller-in-the-loop} dynamics simulation, which rejects 73\% of kinematically valid grasps that would destabilize the controller;
and \textbf{(3)~whole-body affordance-aware alignment}, which samples $K$ diverse candidates and repositions the whole body to the configuration of highest \emph{execution} quality rather than highest grasp confidence.
Trained entirely from synthetic data with zero real demonstrations, our system attains 87.5\% real-world success on a multi-level shelf task, generalizes across novel objects (90/85\% ID/OOD), holds $\ge$73\% at every tested height, and transfers zero-shot to diverse everyday scenes with cluttered backgrounds and unseen objects. At scale in simulation, the same policy extrapolates to table heights well below its training range, where every single-component ablation loses 25--81 points: the learned affordances are grounded in object geometry and whole-body feasibility, not in memorized configurations.

Our contributions are:
\begin{itemize}[leftmargin=*, itemsep=1pt, topsep=2pt, partopsep=0pt, parsep=0pt]
    \item \textbf{Kinodynamic-aware whole-body affordance learning.} A large-scale synthetic data engine that co-optimizes whole-body posture with the hand and enforces controller-in-the-loop dynamic feasibility, so the learned affordances are kinodynamically feasible by construction.
    \item \textbf{A zero-demonstration Perceive--Align--Grasp system.} A complete humanoid pipeline that discovers occluded targets, repositions the whole body to an executable grasp configuration, and executes it under whole-body control, without any real-world demonstration.
    \item \textbf{Extensive real-world validation.} The complete pipeline runs zero-shot on a Unitree G1 humanoid, outperforming imitation and grasp-planning baselines and generalizing to unseen objects, heights, and everyday scenes.
\end{itemize}
\secv
\section{Related Work}

\ssecv
\subsection{Humanoid Whole-Body Loco-Manipulation}
\ssecv
Coordinating locomotion and manipulation on humanoids couples floating-base stability with end-effector precision. Learned whole-body controllers now track expressive motions and coordinate legs and arms under contact and force disturbances~\citep{cheng2024expressive,lu2024mobiletelevision,liao2025beyondmimic,yang2025omniretarget,zhang2025falcon,sun2025ulc,wei2025hmc,chen2025chip,10802754,niu2026touchdream}, and teleoperation systems built on them have made whole-body demonstration collection practical enough to train capable manipulation policies~\citep{ben2025homiehumanoidlocomanipulationisomorphic,ze2025twist2scalableportableholistic,he2024learning,ji2024exbody2,luo2025sonic,li2025cloneclosedloopwholebodyhumanoid,li2025amoadaptivemotionoptimization,he2024omnih2o,fu2024humanplus,ze2025twist,cheng2024opentelevision,li2026omniclone,zhu2026clot,nai2026hmi,heng2026humdex,li2024okami,qiu2025humanoidpolicy,ze2024idp3,niu2026touchdream,jiang2026crosshandlatentrepresentationvisionlanguageaction,qiu-song-peng-2024-wildlma}. These policies, however, remain bounded by their demonstration coverage, and on a floating base, pose drift compounds over long-horizon rollouts. Sim-to-real visual policies~\citep{he2025viral,xue2025openingsimtorealdoorhumanoid,liu2024visual,lin2025sim,xu2024humanvla,wei2026psi0openfoundationmodel,luo2024omnigrasp,luo2025emergent,liu2025fetchbotlearninggeneralizableobject,dong2026herolearninghumanoidendeffector,sun2026dexweave} remove the demonstration bottleneck, most directly VIRAL~\citep{he2025viral}, but leave perception to data scaling. In contrast, we formulate perception as an active optimization problem, coordinating whole-body motion to search for and maximize target observability.

\ssecv
\subsection{Learning from Synthetic Data}
\ssecv
The cost of real interaction data has driven a shift toward simulation-based learning, from domain randomization~\citep{tobin2017domainrandomizationtransferringdeep,Peng_2018,openai2019learningdexterousinhandmanipulation,openai2019solvingrubikscuberobot,tan2018simtoreallearningagilelocomotion,kumar2021rmarapidmotoradaptation,xie2018feedbackcontrolcassiedeep,wei2025ensuringforcesafetyvisionguided} and synthetic grasp datasets~\citep{ye2025dex1b,wang2023dexgraspnet,zhang2024dexgraspnet20learninggenerative,xu2023unidexgrasp,fang2020graspnet,mahler2017dexnet20deeplearning,zhang2026unidex,chao2021dexycb,ye2025powerprecisionlearningfinegrained} to generative, automated pipelines~\citep{wang2024robogenunleashinginfinitedata,mandlekar2023mimicgendatagenerationscalable,xue2025demogensyntheticdemonstrationgeneration,wang2024gensimgeneratingroboticsimulation,yu2023scalingrobotlearningsemantically}. Most directly comparable to ours, billion-scale grasping foundation models~\citep{deng2025graspvlagraspingfoundationmodel} backed by GPU-accelerated simulators~\citep{nvidia2025isaaclabgpuacceleratedsimulation,tao2025maniskill3gpuparallelizedrobotics} and large asset libraries~\citep{deitke2022objaverseuniverseannotated3d} achieve zero-shot generalization in cluttered scenes. These efforts, however, predominantly target fixed-base settings or omit the dynamical plausibility of generated samples, admitting candidates that are kinematically valid but dynamically infeasible under closed-loop whole-body control. In contrast, we co-optimize whole-body posture with the hand for each grasp and enforce kinodynamic feasibility under whole-body execution on a floating base, with the deployment controller in the loop.

\ssecv
\subsection{Active Perception for Manipulation}
\ssecv
Active perception spans reconstruction, recognition, and manipulation~\citep{bajcsy1988active,eidenberger2010active,isler2016information,delmerico2018comparison,dengler2023viewpoint,placed2023surveyactivesimultaneouslocalization,zou2025m33dspatialmultimodalmemory}, and has been coupled with grasping on fixed arms and mobile manipulators~\citep{jangir2022look,breyer2022closed,jauhri2024active,hu2025realworldreinforcementlearningactive}, where the camera moves largely independently of balance. On humanoids the camera is body-attached, so a viewpoint change is a whole-body motion~\citep{kim2024armoregocentricperceptionhumanoid,liu2024visual,yin2025womapworldmodel}, and reaching a viewpoint is itself a whole-body planning problem~\citep{kaiser2018affordance}. Recent humanoid systems address this only partially: ViA~\citep{xiong2025via} learns coordinated head--bimanual control from demonstrations but is bounded by their coverage, and Choice Policy~\citep{qi2025coordinated} regulates an active head while leaving the body passive, so the camera cannot reach the configurations occlusion demands.
\secv
\section{Method}
\secv
\label{sec:method}

\begin{figure}[!t]
    \centering
    \includegraphics[width=\linewidth]{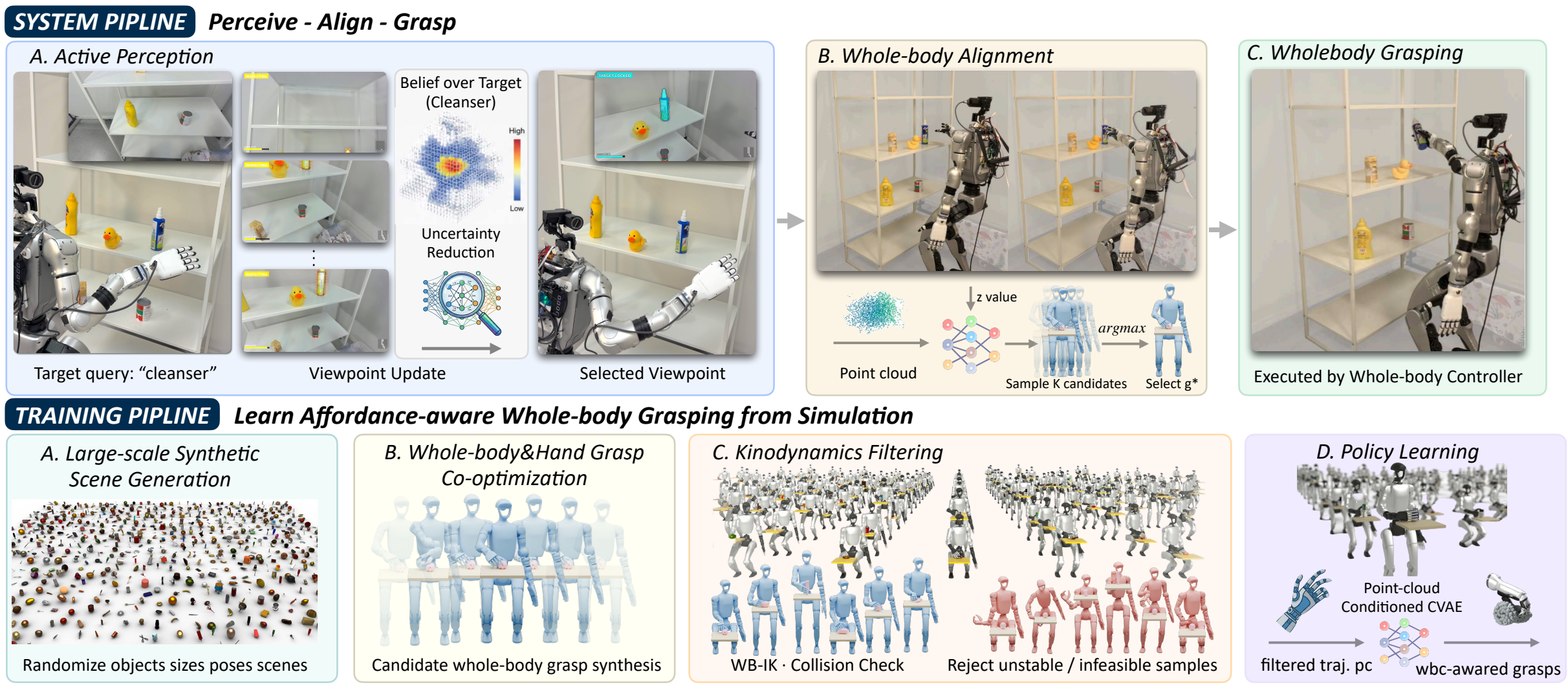}
    \caption{System and training pipelines. The system pipeline (top) follows a Perceive--Align--Grasp structure, combining active perception, whole-body alignment, and robust grasp execution. The training pipeline (bottom) constructs large-scale synthetic data with kinodynamic filtering to learn affordance-aware whole-body grasping.}
    \label{fig:pipeline}
\end{figure}

Our framework comprises two complementary pipelines (Fig.~\ref{fig:pipeline}).
The \emph{system pipeline} follows a Perceive--Align--Grasp structure:
active perception (Sec.~\ref{sec:active_perc}) discovers the target under egocentric occlusion;
the affordance model proposes whole-body grasp configurations, which alignment ranks and repositions the body toward;
and a collision-aware planner executes the selected grasp through the whole-body controller (Sec.~\ref{sec:alignment}).
The \emph{training pipeline} builds large-scale synthetic data filtered for kinodynamic feasibility, yielding the affordance model that drives both alignment and grasping (Sec.~\ref{sec:affordance}).
Both share one hierarchical whole-body controller coupling upper-body IK with an RL lower-body policy (Sec.~\ref{sec:loco_manip}).

\ssecv
\subsection{Hierarchical Whole-Body Controller}
\ssecv
\label{sec:loco_manip}

Whole-body grasping places two competing demands on the controller:
\emph{task-space precision} for end-effector placement, and \emph{floating-base stability} across the diverse postures that different object heights and reaches require.
End-to-end policies struggle to meet both under the tight tolerances of zero-shot sim-to-real transfer.

A hierarchical two-layer design decouples these concerns.
The high-level \textbf{whole-body IK}~\citep{ye2026robokit} maps desired end-effector poses (wrist, hand, and head, in the world frame) to upper-body joint targets $\mathbf{q}_{\text{upper}}$ and lower-body commands $\mathbf{u}_{\text{lower}}=(v_x,v_y,v_{\mathrm{yaw}},h)$, with $h$ the pelvis height.
For the low-level \textbf{RL locomotion policy}, we use the decoupled whole-body controller of SONIC (GR00T)~\citep{luo2025sonic}, which tracks $\mathbf{u}_{\text{lower}}$ from proprioception, outputting lower-body targets that realize the motion while maintaining balance.

This coupling poses a challenge: as the RL policy shifts pelvis height or base position, the root pose drifts in the world frame, and an IK solver blind to this drift renders world-frame targets unreachable---fatal for centimeter-level grasping.
A closed-loop \emph{pelvis compensator} closes the gap, pre-warping every task-space target by the low-pass-filtered mismatch between the IK-assumed and FK-realized pelvis pose (App.~\ref{app:wbc} gives the $SE(3)$ update and filter).
Crucially, the \emph{same} controller drives synthetic-data filtering (Sec.~\ref{sec:affordance}), so training data is supported on actions the deployment system can actually execute.

\ssecv
\subsection{Active Perception}
\ssecv
\label{sec:active_perc}

We cast object discovery under egocentric occlusion as sequential viewpoint selection that maximizes the probability of detecting the target.
The module is a finite-state machine with a \emph{Search} mode (global exploration) and an \emph{Inspect} mode (local refinement after detection); the target is named by a text query to Grounding DINO~\citep{liu2024groundingdino} (e.g., ``red cup''), making acquisition language-conditioned without a separate language module.

\noindent\textbf{Search.}
We maintain a normalized belief $p(i,j)$, the probability that the target is visible from camera pose $(\mathrm{yaw}_i,z_j)$, over a discretized grid of camera yaw and height.
Each camera pose is a head-pose target for the whole-body controller (Sec.~\ref{sec:loco_manip}), so reaching a viewpoint is a whole-body motion: the robot squats or rises to look, not only turns its head.
For a candidate pose $T$, the view frustum projects to a mask $\mathcal{M}(T)$ with \emph{probability of detection} $\mathrm{PoD}(T)=\sum_{(i,j)\in\mathcal{M}(T)}p(i,j)$.
We execute the viewpoint maximizing $\mathrm{score}(T)=\mathrm{PoD}(T)-w_c(d_{\mathrm{yaw}}^2+d_z^2)-w_o\,r_{\mathrm{overlap}}$, with $(d_{\mathrm{yaw}},d_z)$ the displacement from the current camera pose, trading detection likelihood against camera motion and revisiting.

\noindent\textbf{Belief update.}
After observing Grounding DINO confidence $s_t$ at the executed viewpoint $T_t$, cells in the frustum $\mathcal{M}_t$ are updated by a Bayes-style multiplicative rule,
\begin{equation}
p_{t+1}(i,j)\;\propto\;\lambda_t(i,j)\,p_t(i,j),
\qquad \textstyle\sum_{i,j}p_{t+1}=1,
\end{equation}
with a detection-dependent likelihood ratio $\lambda_t$ that \emph{amplifies} frustum cells on a hit ($\lambda_t=1+\alpha\,\tfrac{s_t-\tau}{1-\tau}$ for $s_t\!\ge\!\tau$, rising with confidence above threshold $\tau$), \emph{decays} them on a miss ($\lambda_t=1-\gamma$ for $s_t\!<\!\tau$), and leaves cells outside the frustum unchanged.
Hits concentrate belief; misses redistribute mass to unvisited viewpoints, driving systematic exploration.

\noindent\textbf{Inspect and termination.}
Search hands off to Inspect once $\max_{i,j}p(i,j)$ stays above a convergence threshold for $k$ steps; Inspect then refines the viewpoint by point-cloud visibility quality (visible-point count, depth continuity, normal stability) and terminates when observation quality saturates, passing a stable observation to alignment.

\ssecv
\subsection{Learning Kinodynamic-Aware Whole-Body Affordance}
\ssecv
\label{sec:affordance}

Given a stable observation, the robot must decide \emph{how} to approach and grasp the object with its whole body.
We learn a \emph{whole-body affordance policy} that maps a single-view segmented point cloud $\mathbf{P}$ to candidate \emph{keyframe actions}, each a (pre-grasp, grasp, post-grasp) sequence of end-effector poses and hand commands executed by the whole-body controller.
Every training sample has survived controller-in-the-loop execution, so the learned affordances are kinodynamically feasible by construction.

\noindent\textbf{Large-scale World-frame Randomization.}
Unlike fixed-base grasping, where object variation is largely in-plane, a floating-base humanoid must adapt its \emph{entire body} to different object heights and reaches.
We randomize table geometry and object placement across a wide reachable workspace in the world frame, with objects drawn from a large-scale mesh dataset under randomized scale and mass (App.~\ref{app:datagen}).
This forces diverse whole-body strategies: squatting for low objects, extending for far ones, and adapting the whole-body configuration to everything in between.

\noindent\textbf{Kinodynamic-aware Data Engine.}
We build the dataset through a three-stage pipeline:
(i)~\emph{whole-body--hand grasp co-optimization} jointly synthesizes the whole-body configuration and hand joints for each of $N_{\mathrm{cand}}$ candidates by energy-based optimization in joint space~\citep{ye2026robokit}, coupling hand contact with a reachable whole-body posture, unlike gripper-based or fixed-upper-body generators that ignore humanoid morphology (energy terms in App.~\ref{app:datagen});
(ii)~\emph{motion planning} builds collision-free pre-grasp$\to$grasp$\to$post-grasp trajectories;
(iii)~\emph{controller-in-the-loop dynamics simulation} rolls out each trajectory under the same WB-IK\,+\,RL stack used at deployment.
A trajectory is kept only if it satisfies three criteria: \emph{task success} (object lifted $\ge$5\,cm and held), \emph{stable execution} (no fall, bounded base attitude), and \emph{trackability} (end-effector error within deployment tolerances).
This last stage is the key to kinodynamic awareness: in our profiling it rejects 73\% of kinematically valid grasps (Sec.~\ref{sec:kinodyn}), because the whole-body controller cannot stay stable through the reach or IK tracking diverges under interaction forces.
We further filter by camera visibility to avoid degenerate observations (App.~\ref{app:datagen}).

\noindent\textbf{Observation and Action Spaces.}
The \emph{observation} is a segmented single-view partial point cloud $\mathbf{P}\in\mathbb{R}^{N\times3}$ of the target, transformed from camera to robot task frame to match the egocentric deployment view.
The \emph{action} is a three-stage keyframe trajectory $\mathbf{a}=(\mathbf{a}^{\mathrm{pre}},\mathbf{a}^{\mathrm{grasp}},\mathbf{a}^{\mathrm{post}})$, where each stage $\mathbf{a}^{s}=[\,\mathbf{p}^{s},\mathbf{r}^{s},\mathbf{h}^{s}\,]$ gives end-effector translation $\mathbf{p}^{s}\in\mathbb{R}^{3}$, orientation $\mathbf{r}^{s}\in\mathbb{S}^{3}$ (unit quaternion), and hand joint targets $\mathbf{h}^{s}\in\mathbb{R}^{6}$.

\noindent\textbf{Whole-body Affordance Policy.}
Since a point cloud admits multiple feasible grasps, we model the affordance as a \emph{multimodal} distribution with a point-cloud conditioned CVAE: a PointNet~\citep{qi2017pointnet} encodes $\mathbf{P}$, conditioning a CVAE that decodes the keyframe action from a sampled latent.
At inference we draw several latents for diverse proposals, ranked and executed by the alignment stage (Sec.~\ref{sec:alignment}; architecture and training in App.~\ref{app:policy}).

\ssecv
\subsection{Whole-Body Alignment and Execution}
\ssecv
\label{sec:alignment}

Once active perception settles on a stable viewpoint, the target is segmented with SAM~\citep{kirillov2023segany} into an object-only point cloud and passed to the affordance model (Sec.~\ref{sec:affordance}) for grasp proposal and whole-body alignment.

\noindent\textbf{Candidate Generation and Ranking.}
The CVAE decoder draws $K$ latents into $K$ diverse candidate keyframe sequences $\{\mathbf{a}_k\}_{k=1}^{K}$; trained only on kinodynamically verified data, its proposals are strongly biased toward configurations the whole-body controller can execute.
For each, we solve whole-body IK for the pre-grasp target and rank by $\mathrm{rank}(\mathbf{a}_k)=w_{\mathrm{ik}}\,Q_{\mathrm{IK}}(\mathbf{a}_k)-w_{\mathrm{prox}}\,\|\mathbf{q}_k-\mathbf{q}_{\mathrm{curr}}\|$, where $Q_{\mathrm{IK}}$ is IK quality (inverse residual norm) and $\|\mathbf{q}_k-\mathbf{q}_{\mathrm{curr}}\|$ the joint-space reconfiguration cost.
This favors kinematically achievable proposals that need minimal whole-body motion, which matters for floating-base stability.
IK evaluation is batched and lightweight (${\sim}1$\,ms for all $K$ candidates).

\noindent\textbf{Execution.}
The top-ranked candidate is executed via collision-aware Cartesian motion planning through the whole-body controller (Sec.~\ref{sec:loco_manip}); if planning fails, e.g., due to scene collision, the system falls back to the next candidate in ranked order.
The robot approaches the pre-grasp pose, closes the hand, and performs the post-grasp lift.
Unlike methods that select from a predefined library of stable poses, our alignment configurations are \emph{learned} from kinodynamically filtered data, \emph{object-specific} via point-cloud conditioning, and \emph{dynamics-aware} by construction.

\secv
\section{Experiments}
\secv
\label{sec:experiments}

We evaluate our system through three complementary evaluations:
Task~1 tests full pipeline integration under realistic shelf conditions;
Task~2 isolates manipulation from perception with oracle observations on real hardware;
Task~3 measures spatial and object generalization at scale in simulation, including \emph{extrapolation beyond the training height range}.
Beyond aggregate comparison, we investigate whether large-scale kinodynamically filtered synthetic training generalizes across the full humanoid workspace and novel objects, which pipeline component fails in which regime, and how performance scales with synthetic data volume.

\ssecv
\subsection{Experimental Setup}
\label{sec:exp_setup}

\textbf{Hardware.}
We evaluate on the Unitree G1 humanoid with a 3-DOF active head~\citep{li2025amoadaptivemotionoptimization}, ZED Mini stereo camera, and Inspire dexterous hands.
All inference runs on an external desktop (RTX~5090); commands are transmitted via Ethernet to the onboard computer.

\textbf{Task~1: Full Pipeline} (\emph{Perceive\,$\to$\,Align\,$\to$\,Grasp}).
Objects are placed on a multi-level shelf with partial occlusions and randomized heights.
The robot starts from a nominal pose with no prior knowledge of the target and must autonomously discover, align to, and grasp the object.
We run $N{=}40$ trials over 10 YCB~\citep{calli2015ycb} objects (5\,ID, 5\,OOD); OOD objects are held out from the training set.
Target objects are specified via natural language, conditioning the DINO text prompt for detection.

\textbf{Task~2: Align \& Grasp} (\emph{Oracle Perception}).
To isolate manipulation from perception confounds, we provide oracle observations on a height-adjustable table at five controlled levels:
H1\,(0.50\,m), H2\,(0.60\,m), H3\,(0.70\,m), H4\,(0.80\,m), H5\,(0.85\,m);
10 objects $\times$ 5 heights $\times$ 3 trials\,=\,150 (75\,ID, 75\,OOD).
Task~2 is the hardware anchor for the height profile measured at scale in Task~3.
ACT and ViA receive demonstrations at H2 and H4 only, so the remaining heights measure their sensitivity to demonstration \emph{coverage}; our method trains on continuously randomized heights in simulation.

\textbf{Task~3: Large-Scale Simulation Evaluation.}
To evaluate spatial generalization at a scale impractical on hardware, we sweep
17 table heights (0.10--0.90\,m in 5\,cm steps) with randomized planar
placement and randomized object yaw,
over two object catalogs:
\textbf{YCB} (12 everyday objects) and \textbf{RoboCasa}~\citep{nasiriany2024robocasa}
(11 objects spanning a broader shape distribution: smaller, thinner, and more
elongated geometries, e.g., peeler, whisk, chicken drumstick).
Six system variants (Ours and five ablations/baselines, below) are evaluated on
\emph{identical} scene sets, so every delta is paired; ${\sim}$18{,}800 trials in
total. Success is scored from telemetry under a criterion \emph{stricter} than the
real-world one: the object must be lifted $\ge$5\,cm and held for 2\,s with bounded
in-hand slip, and the robot must remain upright (App.~\ref{sec:supp_sim_eval}).
Crucially, training tables span only 0.40--0.90\,m: all evaluations at
0.10--0.35\,m probe \emph{extrapolation below the training range}.

\textbf{Metrics.}
Success rate~(SR): object lifted ${\geq}5$\,cm and stably held.
For Task~1 we additionally report \emph{Det.}\,(\%): object successfully localized, and \emph{Gr$|$Det}\,(\%): grasp success given detection, which decomposes failures into perception vs.\ manipulation stages.

\textbf{Baselines.}
We compare against end-to-end imitation methods, modular grasp planners, and ablations of our pipeline.
\emph{Vanilla ACT}~\citep{zhao2023learning}: end-to-end visuomotor policy with a fixed head and no active viewpoint control.
\emph{ACT\,+\,Active Head}: ACT with the active head enabled during both demonstration collection and policy execution, isolating the contribution of active perception.
\emph{ViA}~\citep{xiong2025via}: diffusion policy with learned active perception, adapted to floating-base control.
\emph{GraspNet\,+\,WB-IK}~\citep{fang2020graspnet}: parallel-jaw grasp poses from the pretrained checkpoint, mapped to dexterous-hand pre-grasps and executed via our WB-IK controller (\emph{+\,AP}: adding our active perception).
\emph{Dex1B (adapted)\,+\,WB-IK\,+\,AP}~\citep{ye2025dex1b}: large-scale dexterous grasp framework adapted to our humanoid.
Data is generated across the full workspace with upper-body optimization (arm IK only, no whole-body--hand co-optimization), without dynamics filtering or alignment.
Dex1B shares our CVAE\,+\,PointNet architecture and hand type, directly isolating the contribution of kinodynamic filtering and whole-body alignment.
ACT and ViA are trained on teleoperated demonstrations of ID objects
(Task~1: 70 full-pipeline demonstrations per object at every evaluated shelf level, with randomized horizontal placement, so their height coverage matches the test conditions; Task~2: grasp-only at H2/H4).
Our method uses zero real demonstrations: 600k synthetic trajectories only.

\textbf{Ablations.}
\emph{--Perceive}: fixed camera pose, bypassing active viewpoint search.
\emph{--Whole-Body}: active perception is enabled for object discovery, but the base is fixed during grasping, with no squatting or torso adaptation (arm-only manipulation).
\emph{--Align}: skip the whole-body alignment stage; instead of ranking $K$ candidates and repositioning the body, sample a single grasp candidate and execute directly from the current configuration.
\emph{--Dyn.\ Filter}: remove dynamics simulation from the data pipeline, retaining only kinematic and collision filtering.
\emph{--Compensator}: disable the closed-loop pelvis compensator (Sec.~\ref{sec:loco_manip}), so task-space targets are not corrected for pelvis drift.

\begin{figure}[t]
\centering
\captionsetup{font=footnotesize}
\captionsetup[sub]{skip=1pt}
\renewcommand{\arraystretch}{0.9}
\begin{subfigure}[t]{0.47\linewidth}
\vspace{0pt}
\centering
\scriptsize
\setlength{\tabcolsep}{3pt}
\begin{tabular}{llcccc}
\toprule
& \multirow{2}{*}{\textbf{Method}} &
\multicolumn{2}{c}{\textbf{SR (\%)}} &
\multicolumn{2}{c}{\textbf{Stage (\%)}} \\
\cmidrule(lr){3-4}\cmidrule(lr){5-6}
& & All & ID/OOD & Det. & Gr$|$Det \\
\midrule
\multirow{6}{*}{\rotatebox{90}{\scriptsize Baselines}}
& Vanilla ACT              & 32.5 & 55/10  & 52.5 & 61.9 \\
& ACT + Active Head        & 40.0 & 65/15  & 57.5 & 69.6 \\
& ViA (adapted)            & 50.0 & 85/15  & 70.0 & 71.4 \\
& GraspNet + WB-IK         & 22.5 & --     & 50.0 & 45.0 \\
& \;\;+ Our Perc.          & 40.0 & --     & 90.0 & 44.4 \\
& Dex1B + WB-IK + AP       & 52.5 & 55/50  & 92.5 & 56.8 \\
\midrule
& \textbf{Ours (Full)}     & \textbf{87.5} & \textbf{90/85} & \textbf{92.5} & \textbf{94.6} \\
\midrule
\multirow{4}{*}{\rotatebox{90}{\scriptsize Abl.}}
& --Perceive               & 50.0 & 55/45  & 57.5 & 87.0 \\
& --Whole-Body             & 20.0 & 20/20  & 87.5 & 22.9 \\
& --Align                  & 62.5 & 65/60  & 85.0 & 73.5 \\
& --Dyn.\ Filter           & 62.5 & 65/60  & 90.0 & 69.4 \\
\bottomrule
\end{tabular}
\caption{}
\label{tab:task1}
\end{subfigure}
\hfill
\begin{minipage}[t]{0.50\linewidth}
\vspace{0pt}
\begin{subfigure}{\linewidth}
\centering
\includegraphics[width=\linewidth]{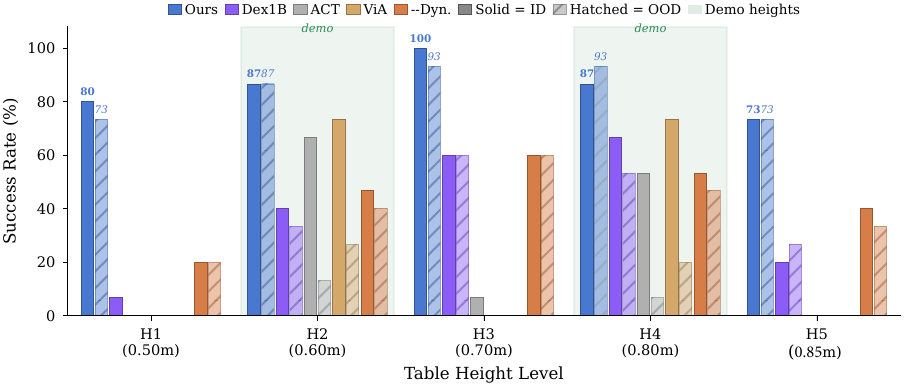}
\caption{}
\label{fig:spatial}
\end{subfigure}

\vspace{3pt}
\begin{subfigure}{\linewidth}
\centering
\scriptsize
\setlength{\tabcolsep}{3pt}
\begin{tabular}{clrr}
\toprule
\textbf{Stage} & \textbf{Filter} & \textbf{Output} & \textbf{Survival} \\
\midrule
1 & Grasp opt.\ + IK + coll.  & 12.4\,M & 2.4\% \\
2 & Collision-free MP         & 298\,K  & 2.4\% \\
3 & Dynamics sim.\ (WBC)      & 80.7\,K & 27.0\% \\
\midrule
  & Final (+ view sel.)       & 18\,K   & --- \\
\bottomrule
\end{tabular}
\caption{}
\label{tab:filtering}
\end{subfigure}
\end{minipage}
\caption{\textbf{Quantitative results.}
\textbf{(a)}~Task~1 full-pipeline evaluation ($N{=}40$, 10 objects, 5\,ID/5\,OOD); \emph{Det.}~= object localized, \emph{Gr$|$Det}~= grasp success given detection.
\textbf{(b)}~Task~2 spatial generalization across five heights (oracle perception, 10 objects; solid = ID, hatched = OOD, green = ACT/ViA demo heights): ours $\ge$73\% everywhere, while ACT/ViA succeed only at their demonstrated heights (a demonstration-coverage effect; Sec.~\ref{sec:spatial}) and Dex1B/--Dyn.\ Filter degrade at the extremes.
\textbf{(c)}~Kinodynamic data filtering: 73\% of kinematically valid grasps fail under the deployment whole-body controller (508\,M SDF candidates $\to$ 18\,K training grasps).}
\label{fig:results}
\vspace{-12pt}
\end{figure}

\ssecv
\subsection{Main Results}
\ssecv
\label{sec:main_results}

Fig.~\ref{tab:task1} summarizes Task~1. Our full system reaches 87.5\% SR (90/85 ID/OOD) with 94.6\% Gr$|$Det (once the object is found, grasping almost always succeeds), using zero real demonstrations, while the strong imitation baseline (ViA) needs teleoperated demos to reach 50\%.

\textbf{Baselines fail along distinct axes.}
\emph{End-to-end imitation} memorizes object-specific behavior: ViA holds 85\% ID but collapses to 15\% OOD, as cumulative errors over the long-horizon floating-base action space stall it on unseen objects.
\emph{Gripper-based planners} fail at execution, not discovery: adding our perception to GraspNet lifts Det.\ 50$\to$90\%, yet Gr$|$Det is unchanged (45.0$\to$44.4\%), because parallel-jaw pre-grasps give postures unnatural for the humanoid hand (Fig.~\ref{fig:supp_failure}c) and hand--environment collisions (Fig.~\ref{fig:supp_failure}d).
\emph{Large-scale dexterous data alone} is insufficient: Dex1B (adapted), which shares our CVAE\,+\,PointNet and hand type, matches our Det.\ (92.5\%) yet reaches only 56.8\% Gr$|$Det, below \emph{--Dyn.\ Filter} (69.4\%): built for tabletop settings, its data lacks both whole-body--hand co-optimization and dynamic feasibility (Sec.~\ref{sec:signatures} dissects this by height).

\textbf{Each module addresses a distinct failure stage.}
Stage-wise metrics separate perception (Det.) from manipulation (Gr$|$Det): \emph{--Perceive} cuts Det.\ by 35\,pts (perception is the discovery bottleneck, Sec.~\ref{sec:active_perc_exp}), while the three manipulation ablations each collapse Gr$|$Det along a different axis.
\emph{--Whole-Body} falls to 22.9\% (cannot reach without squatting), \emph{--Align} to 73.5\% (sub-optimal approach from the current pose), and \emph{--Dyn.\ Filter} to 69.4\% (dynamically infeasible training data). The full 94.6\% confirms all four are necessary; we dissect them next and resolve each ablation's failure \emph{regime} across the workspace in Sec.~\ref{sec:signatures}.

\ssecv
\subsection{Analysis}
\ssecv
\label{sec:analysis}

\subsubsection{Kinodynamic filtering rejects 73\% of valid grasps}
\label{sec:kinodyn}

\begin{figure}[t]
\centering
\captionsetup{font=footnotesize}
\begin{subfigure}{\linewidth}
    \centering
    \includegraphics[width=0.9\linewidth]{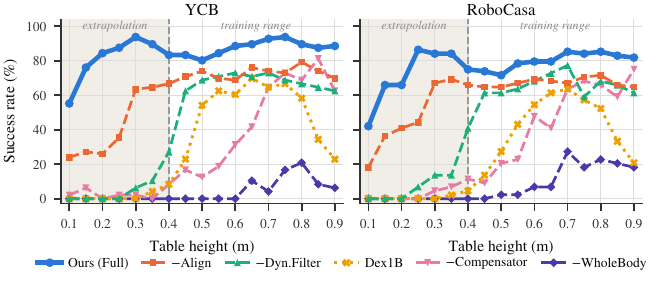}
    \caption{}
    \label{fig:sim_eval_curves}
\end{subfigure}

\vspace{2pt}
\begin{subfigure}{0.72\linewidth}
    \centering
    \scriptsize
    \setlength{\tabcolsep}{4pt}
    \renewcommand{\arraystretch}{0.95}
    \begin{tabular}{@{}lcccc@{}}
    \toprule
    & \multicolumn{2}{c}{\textbf{YCB}} & \multicolumn{2}{c}{\textbf{RoboCasa}} \\
    \cmidrule(lr){2-3}\cmidrule(lr){4-5}
    \textbf{Method} & Extrap. & In-range & Extrap. & In-range \\
    & \scriptsize 0.10--0.35\,m & \scriptsize 0.40--0.90\,m & \scriptsize 0.10--0.35\,m & \scriptsize 0.40--0.90\,m \\
    \midrule
    \textbf{Ours (Full)} & \textbf{81.1} & \textbf{87.4} & \textbf{71.4} & \textbf{79.8} \\
    \midrule
    --Align & 40.1 & 72.4 & 46.0 & 67.2 \\
    --Dyn.\ Filter & 2.8 & 64.4 & 5.7 & 63.6 \\
    Dex1B~+~WB-IK & 0.7 & 47.7 & 0.4 & 39.2 \\
    --Compensator & 2.1 & 43.6 & 1.9 & 44.0 \\
    --Whole-Body & 0.0 & 6.1 & 0.0 & 11.4 \\
    \bottomrule
    \end{tabular}
    \caption{}
    \label{tab:sim_eval_bands}
\end{subfigure}
\caption{\textbf{Large-scale simulation evaluation} (${\sim}$18{,}800 paired trials; Task~3 protocol in Sec.~\ref{sec:exp_setup}).
\textbf{(a)}~SR vs.\ table height on YCB (left) and RoboCasa (right). Training tables span 0.40--0.90\,m (dashed line); the shaded band is \emph{extrapolation below the training range}.
\textbf{(b)}~SR (\%) aggregated over the extrapolation and in-range bands (95\% Wilson CIs for the Full rows are within $\pm 4$\,pts, e.g.\ YCB extrapolation $81.1\,[77.7, 84.1]$).
The full system loses only 6--8\,pts below the training floor; removing any single component collapses extrapolation to 0--6\% (--Align to ${\sim}$40--46\%).}
\label{fig:sim_eval}
\vspace{-12pt}
\end{figure}

Kinematic feasibility alone is insufficient for a floating base. Fig.~\ref{tab:filtering} traces the pipeline on an example profiling run: 508M candidates yield 18K training grasps, and 73\% of kinematically valid trajectories are dynamically infeasible: they satisfy IK, collision, and planning constraints yet jitter (Fig.~\ref{fig:supp_failure}a), twist (Fig.~\ref{fig:supp_failure}b), or destabilize under the deployment controller. Removing the filter (\emph{--Dyn.\ Filter}) drops Gr$|$Det 94.6$\to$69.4\% at near-identical Det.\ (90.0 vs.\ 92.5\%); Dex1B, which also lacks it, drops further (56.8\%); controller-in-the-loop verification is essential for training-data quality.
The consequence is sharpest at scale (Fig.~\ref{fig:sim_eval}): trained on the \emph{same} scenes without the filter, the policy is merely weaker inside the training range (64 vs.\ 87/80\% on YCB/RoboCasa) but collapses to 3--6\% below it---the filter is what turns in-distribution competence into extrapolation (Sec.~\ref{sec:spatial}).

\subsubsection{Extrapolation beyond the training height range}
\label{sec:spatial}

At scale in simulation (Task~3, Fig.~\ref{fig:sim_eval}, ${\sim}$18.8K paired trials), the full system holds 87.4/79.8\% (YCB/RoboCasa) within the trained 0.40--0.90\,m range and 81.1/71.4\% at 0.10--0.35\,m, below the training floor and reached from deep squats at shin height.
It loses only 6--8\,pts under extrapolation, while the other variants collapse there to 0--6\% and \emph{--Align} drops to 40--46\%.
Trained on large-scale, kinodynamically filtered data, the affordance policy learns rich local geometric features under a strong kinodynamic prior: its proposals remain object-relative and controller-executable even when the body configuration leaves the training distribution.
Below the training floor (0.10--0.35\,m), whole-body control and dynamics filtering become indispensable (removing either collapses success to near zero), and alignment's contribution roughly doubles (13--15\,$\to$\,25--41\,pts), as maintaining stable, feasible whole-body execution becomes the dominant factor (Sec.~\ref{sec:signatures}).
Object difficulty and the spatial envelope are analyzed in App.~\ref{sec:supp_sim_eval}.

The full system's profile carries over to hardware. Across five real heights (Task~2, Fig.~\ref{fig:spatial}, 150 trials), ours holds 73--97\% with near-identical ID/OOD (85.3 vs.\ 84.0\%), comparable to the simulated 80--93\% (YCB) at the same heights; Dex1B and \emph{--Dyn.\ Filter} degrade at the extremes, qualitatively consistent with the simulation results.
ACT and ViA, given demonstrations at two heights, collapse at the unseen ones, as expected: teleoperated policies are bounded by their demonstration coverage, and collecting demonstrations across the full workspace is prohibitively expensive---precisely where sim-to-real training on large-scale synthetic data shines.

\subsubsection{Each component has a distinct failure signature}
\label{sec:signatures}

The per-height curves (Fig.~\ref{fig:sim_eval}a) resolve \emph{where} each component matters, which a single ablation number cannot.
\emph{--Compensator} rises with height, from ${\sim}$0\% below 0.40\,m to 59--81\% at 0.70--0.90\,m: pelvis drift grows with squat depth, so uncompensated targets miss most exactly where posture deviates most from nominal; the curve is a direct measurement of drift versus depth.
\emph{--Dyn.\ Filter} shows a cliff at the training floor (0\% at $\le$0.20\,m, $\ge$60\% from 0.45\,m): unfiltered data yields grasps executable only in nominal postures.
\emph{--Whole-Body} is near zero below 0.65\,m and peaks at 21--27\%: this is the arm-only reach envelope, and it quantifies how much of the workspace exists only through the torso and legs.
\emph{--Align} costs a uniform 13--15\,pts inside the training range but 25--41\,pts below it: candidate selection matters most where feasible proposals are scarce.
\emph{Dex1B} is 0\% at $\le$0.30\,m and falls off again above 0.80\,m (34--23\% on YCB): fixed-upper-body generation contains neither low postures nor extended high reaches, a coverage limit of the generator.
Coverage is not the whole gap: even in-range it stays ${\sim}$40\,pts below Full (47.7/39.2 vs.\ 87.4/79.8\%), the compound cost of generating without whole-body--hand co-optimization or dynamic-feasibility filtering.
Together these signatures explain the Task~1 stage-wise ablations (Sec.~\ref{sec:main_results}) mechanistically: each module removes a different failure regime, and only their union covers the workspace.

\begin{figure}[t]
    \centering
    \includegraphics[width=\linewidth]{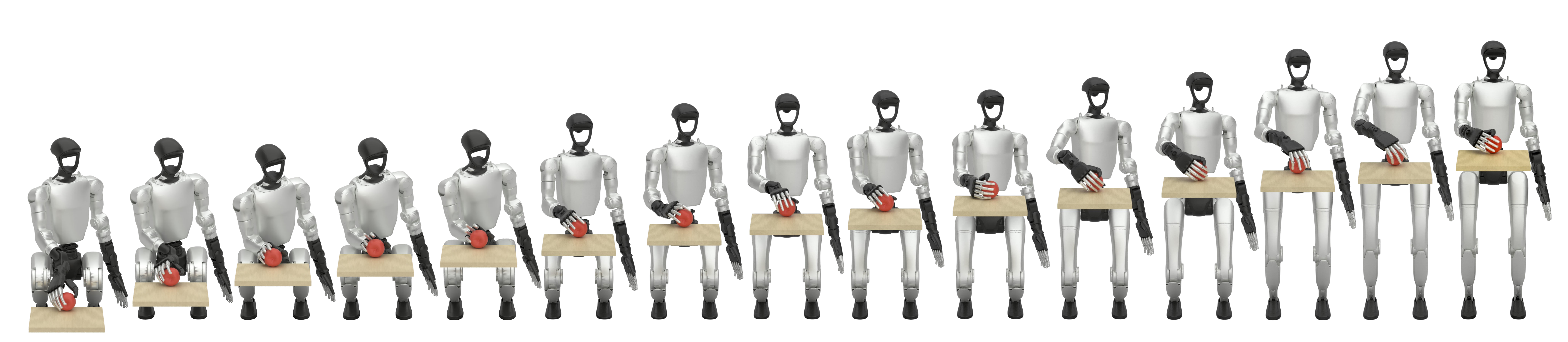}
    \vspace{-18pt}
    \caption{\textbf{Whole-body spatial coverage.} Adapting its entire body, our policy grasps in simulation across a continuous table-height range (0.10--0.90\,m), from deep squats for low targets to full extension for high ones. Training tables spanned 0.40--0.90\,m; the leftmost frames lie below the training range.}
    \label{fig:workspace}
    \vspace{-12pt}
\end{figure}

\subsubsection{Active perception resolves the discovery bottleneck}
\label{sec:active_perc_exp}

Is the bottleneck finding the object or grasping it? \emph{--Perceive} cuts Det.\ by 35\,pts (92.5$\to$57.5\%) while Gr$|$Det stays high (87.0 vs.\ 94.6\%): grasping is reliable given a good view; discovery is the bottleneck. Learned alternatives confirm the gap (ViA 70.0\%, ACT+Active Head 57.5\% Det.\ vs.\ our 92.5\%). Our systematic viewpoint search takes 10--15 updates (${\sim}$8\,s), a long search that end-to-end policies would have to learn jointly with manipulation; this motivates the modular separation of discovery from execution.

\subsubsection{Synthetic data scaling drives generalization}
\label{sec:scaling}

In a dedicated real-world scaling study (App.~\ref{sec:supp_scaling}), SR rises from 9.2\% (5k trajectories) to ${\sim}$89\% at 400k and plateaus thereafter (88.3\% at 600k), with OOD benefiting most (5.0$\to$90.0\%) and the relative OOD degradation narrowing from 62\% to under 2\% above 300k; data diversity progressively closes the generalization gap to novel objects. Scale and filtering are complementary: scale closes the object gap, while kinodynamic filtering is what enables extrapolation beyond the training range (Sec.~\ref{sec:spatial}).

\secv
\section{Conclusion and Limitations}
\secv
\label{sec:conclusion}

We presented a hierarchical Perceive--Align--Grasp framework for humanoid whole-body affordance-aware grasping with active perception. By coordinating belief-driven viewpoint regulation, whole-body alignment, and a kinodynamically filtered affordance model, our system reaches 87.5\% real-world success with zero real demonstrations, outperforming end-to-end imitation and modular grasp-planning baselines by large margins. Three findings anchor the design: 73\% of kinematically valid grasps are dynamically infeasible under the deployment controller, so controller-in-the-loop data curation is essential for floating-base manipulation; perception is the dominant bottleneck under egocentric occlusion (without active perception, detection falls to 57.5\% while Gr$|$Det remains 87\%); and at scale in simulation the policy extrapolates to table heights well below its training range, evidence that kinodynamically filtered affordances are grounded in object geometry and whole-body feasibility rather than memorized configurations.

\textbf{Limitations.} Our system is predominantly \emph{open-loop at the high-level decision stage}, prioritizing interpretability and stability over reactive replanning. The remaining real-world failures (${\sim}$12.5\%) stem mainly from \emph{open-vocabulary detection}, where DINO\,+\,SAM mis-identify targets under severe occlusion or on reflective surfaces, and from occasional collisions at workspace extremes near joint limits. Deployment currently uses double support without stepping, so targets beyond arm-plus-lean reach require repositioning; the controller interface already exposes velocity commands, and extending alignment to stepping is a natural next step.
Synthetic-data generation is also costly (${\sim}$700 GPU-hours per 600k trajectories): fully parallelizable and amortized across deployments, but re-run when the controller or policy architecture changes. Future work may close the loop across perception and execution, co-train the stages end-to-end, and adapt data synthesis to the deployed controller.

\acknowledgments{This work was supported, in part, by NSF CAREER Award IIS-2240014 and NSF CCF-2112665 (TILOS).}

\bibliographystyle{corlabbrvnat}
\bibliography{IEEEabrv, references}

\appendix

\newpage

\section{Hierarchical Whole-Body Loco-Manipulation Controller}
\label{app:wbc}
\label{sec:hier_wbc}

\subsection{Frames and Notation}
\label{sec:frames_notation}

We use the standard rigid-body transform notation ${}^{A}\mathbf{T}_{B}\in SE(3)$, which maps a 3D point expressed in frame $B$ into frame $A$.
The corresponding rotation and translation are ${}^{A}\mathbf{R}_{B}\in SO(3)$ and ${}^{A}\mathbf{p}_{B}\in\mathbb{R}^3$.

\paragraph{Key frames.}
\begin{itemize}[leftmargin=*, itemsep=1pt, topsep=0pt, partopsep=0pt, parsep=0pt]
  \item $\mathcal{W}_0(t)$: the \emph{instantaneous} support frame used by the upper-body controller.
  In double support, its origin is defined at the mid-point of the two contact feet and its yaw is aligned with the average heading of the feet.
  We set roll/pitch to zero to maintain a gravity-aligned frame and to avoid injecting foot tilt noise into the upper-body task frame.
  \item $\mathcal{O}$: the kinematic reference frame of the full-body URDF used for forward kinematics (FK) from joint encoders.
  \item $\mathcal{T}$: pelvis frame, treated as the floating base in the reduced upper-body IK model.
  \item $\mathcal{F}_L,\mathcal{F}_R$: left/right foot frames used as contacts.
\item $\mathcal{E}_L,\mathcal{E}_R$: left/right end-effector frames used by the manipulation tasks.
\item $\mathcal{E}_H$: head camera frame used for active perception.
We treat head gaze as an additional task-space target in the same support frame $\mathcal{W}_0(t)$, i.e.,
${}^{\mathcal{W}_0}\mathbf{T}^{\mathrm{des}}_{\mathcal{E}_H}(t)$ (or equivalently a desired head orientation / optical-axis direction), and include $H$ in the task index set $\mathcal{K}$ when head tracking is enabled.

\end{itemize}

\paragraph{Robot configuration.}
We split the robot state into a lower-body part and an upper-body part:
\begin{equation}
  \mathbf{q} =
  \begin{bmatrix}
    \mathbf{q}_{\text{lower}}\\
    \mathbf{q}_{\text{upper}}
  \end{bmatrix},
\end{equation}
where $\mathbf{q}_{\text{lower}}$ contains leg joints (and any joints actuated by the low-level locomotion controller),
and $\mathbf{q}_{\text{upper}}$ contains waist/torso joints, arms, and optionally head joints (depending on the IK model).

\subsection{Hierarchical Control Decomposition}
\label{sec:hier_decomp}

Our control stack follows a standard outer/inner-loop organization:
(i) an outer-loop whole-body IK module produces task-consistent upper-body commands together with a compact lower-body command interface,
and (ii) an inner-loop RL controller (the decoupled whole-body controller of SONIC (GR00T)~\citep{luo2025sonic}) realizes the commanded pelvis motion while enforcing contact consistency and stability.
A key design requirement is to maintain consistency between the pelvis pose assumed by the reduced IK model and the pelvis motion realized by the low-level controller.

\paragraph{High-level IK (upper-body whole-body IK).}
At time $t$, given desired task-space targets expressed in the support frame $\mathcal{W}_0(t)$,
\begin{equation}
  \Big\{{}^{\mathcal{W}_0}\mathbf{T}^{\text{des}}_{\mathcal{E}_k}(t)\Big\}_{k\in\mathcal{K}},
  \qquad \mathcal{K}\subseteq\{L,R,H\},
\end{equation}
the IK module solves
\begin{equation}
  \big(\mathbf{q}^{\star}_{\text{upper}}(t),\; \mathbf{u}_{\text{lower}}(t)\big)
  =
  \pi_{\text{IK}}\!\Big(
  \{{}^{\mathcal{W}_0}\mathbf{T}^{\text{des}}_{\mathcal{E}_k}(t)\}_{k\in\mathcal{K}},\;
  \widehat{{}^{\mathcal{W}_0}\mathbf{T}_{\mathcal{T}}}(t)
  \Big),
  \label{eq:upper_ik_mapping}
\end{equation}
where $\widehat{{}^{\mathcal{W}_0}\mathbf{T}_{\mathcal{T}}}(t)$ is a contact-consistent reconstruction of the pelvis pose in the same support frame (Sec.~\ref{sec:contact_state_recon}).
This term closes the inter-layer loop by aligning the IK solver's floating-base state with the physically realized pelvis motion.

\paragraph{Low-level RL controller (lower-body tracking).}
The low-level controller executes the command $\mathbf{u}_{\text{lower}}(t)$ and outputs lower-body joint targets:
\begin{equation}
  \mathbf{q}^{\star}_{\text{lower}}(t)
  =
  \pi_{\text{lower}}\!\Big(\mathbf{s}_{\text{lower}}(t),\; \mathbf{u}_{\text{lower}}(t)\Big),
  \label{eq:lower_policy}
\end{equation}
where $\mathbf{s}_{\text{lower}}(t)$ is the proprioceptive observation (joint encoders, IMU, contact flags, etc.).

\paragraph{Lower-body command interface.}
In our standing manipulation experiments (double support, no stepping), we instantiate the general interface $\mathbf{u}_{\text{lower}}=(v_x,v_y,v_{\mathrm{yaw}},h)$ of Sec.~\ref{sec:loco_manip} as a low-dimensional pelvis command, with planar velocities set to zero and yaw commanded as a heading $\psi$ rather than a rate, to reduce inter-layer coupling and improve robustness under rate mismatch and model error:
\begin{equation}
  \mathbf{u}_{\text{lower}}(t) = \big[h(t),\; \psi(t)\big].
  \label{eq:ulower_reduced}
\end{equation}
The same hierarchical formulation directly supports loco-manipulation by extending $\mathbf{u}_{\text{lower}}$ to a more general locomotion interface (e.g., planar velocity and yaw-rate commands) while keeping Eqs.~\eqref{eq:upper_ik_mapping}--\eqref{eq:lower_policy} unchanged.

\paragraph{Commanded output.}
The final motor command is the concatenation of the lower- and upper-body references:
\begin{equation}
  \mathbf{q}^{\star}(t)=
  \begin{bmatrix}
    \mathbf{q}^{\star}_{\text{lower}}(t)\\
    \mathbf{q}^{\star}_{\text{upper}}(t)
  \end{bmatrix}.
\end{equation}
In our deployed system, the low-level controller and the upper-body IK run synchronously at the same control rate (50\,Hz), and the robot manager dispatches the fused command $\mathbf{q}^{\star}(t)$ at each cycle.

\subsection{Contact-Consistent Kinematic State Reconstruction}
\label{sec:contact_state_recon}

A central challenge in hierarchical control is that the pelvis pose \emph{assumed} by the reduced IK model can deviate from the pelvis pose \emph{realized} by the low-level controller due to tracking error, compliance, and modeling mismatch.
To close this gap, we compute a contact-consistent reconstruction of the pelvis pose in the support frame $\mathcal{W}_0(t)$ using joint encoders under double-foot contact.
Importantly, we use an \emph{instantaneous} support frame definition (no integration): all quantities are expressed relative to the current mid-feet frame computed from FK.

\paragraph{Foot/pelvis FK in the URDF reference frame.}
Given measured joint positions $\mathbf{q}(t)$, we compute the forward kinematics in the full-body URDF:
\begin{equation}
  {}^{\mathcal{O}}\mathbf{T}_{\mathcal{F}_L}(t),\;
  {}^{\mathcal{O}}\mathbf{T}_{\mathcal{F}_R}(t),\;
  {}^{\mathcal{O}}\mathbf{T}_{\mathcal{T}}(t)
  = \mathrm{FK}\big(\mathbf{q}(t)\big).
  \label{eq:fk_full}
\end{equation}

\paragraph{Instantaneous mid-feet support frame.}
We define $\mathcal{W}_0(t)$ from the two contact feet:
\begin{align}
  {}^{\mathcal{O}}\mathbf{p}_{\mathcal{W}_0}(t)
  &= \tfrac{1}{2}\Big({}^{\mathcal{O}}\mathbf{p}_{\mathcal{F}_L}(t) + {}^{\mathcal{O}}\mathbf{p}_{\mathcal{F}_R}(t)\Big),\\
  \bar{\psi}(t)
  &= \mathrm{avgYaw}\Big({}^{\mathcal{O}}\mathbf{R}_{\mathcal{F}_L}(t),\, {}^{\mathcal{O}}\mathbf{R}_{\mathcal{F}_R}(t)\Big),\\
  {}^{\mathcal{O}}\mathbf{R}_{\mathcal{W}_0}(t)
  &= \mathbf{R}_z\big(\bar{\psi}(t)\big),\\
  {}^{\mathcal{O}}\mathbf{T}_{\mathcal{W}_0}(t)
  &=
  \begin{bmatrix}
    {}^{\mathcal{O}}\mathbf{R}_{\mathcal{W}_0}(t) & {}^{\mathcal{O}}\mathbf{p}_{\mathcal{W}_0}(t)\\
    \mathbf{0}^\top & 1
  \end{bmatrix}.
  \label{eq:w0_from_feet}
\end{align}
This matches the controller's task frame: centered between the feet and aligned with the support yaw, while remaining gravity-aligned.

\paragraph{Pelvis pose in the support frame.}
The reconstructed pelvis pose used by IK is
\begin{equation}
  \widehat{{}^{\mathcal{W}_0}\mathbf{T}_{\mathcal{T}}}(t)
  =
  \Big({}^{\mathcal{O}}\mathbf{T}_{\mathcal{W}_0}(t)\Big)^{-1}\; {}^{\mathcal{O}}\mathbf{T}_{\mathcal{T}}(t).
  \label{eq:torso_in_w0}
\end{equation}

\paragraph{Remark (estimator modularity).}
While we use a purely kinematic, contact-consistent reconstruction in double support for simplicity and low latency, the proposed interface only requires an estimate of the pelvis pose in the task frame, $\widehat{{}^{\mathcal{W}_0}\mathbf{T}_{\mathcal{T}}}(t)$.
In general, $\widehat{{}^{\mathcal{W}_0}\mathbf{T}_{\mathcal{T}}}(t)$ can be provided by any state estimator that is consistent with contact constraints, e.g., IMU--encoder fusion, visual--inertial odometry with contact updates, or factor-graph filtering, without changing the downstream compensation in Sec.~\ref{sec:closed_loop_comp}.

\subsection{Closed-Loop Pelvis Consistency for Upper-Body IK}
\label{sec:closed_loop_comp}

Let ${}^{\mathcal{W}_0}\mathbf{T}^{\text{IK}}_{\mathcal{T}}(t)$ denote the pelvis pose \emph{assumed} by the IK model (i.e., the floating-base state inside the reduced IK solver),
and let $\widehat{{}^{\mathcal{W}_0}\mathbf{T}_{\mathcal{T}}}(t)$ be the contact-consistent reconstruction from Eq.~\eqref{eq:torso_in_w0}.
We compute the mismatch transform
\begin{equation}
  \Delta \mathbf{T}_{\text{pelvis}}(t)
  =
  \widehat{{}^{\mathcal{W}_0}\mathbf{T}_{\mathcal{T}}}(t)\;
  \Big({}^{\mathcal{W}_0}\mathbf{T}^{\text{IK}}_{\mathcal{T}}(t)\Big)^{-1}.
  \label{eq:delta_torso}
\end{equation}

\paragraph{Compensating task targets (equivalently, updating the tracking frame).}
For any desired end-effector target expressed in $\mathcal{W}_0(t)$, we compensate by the inverse mismatch:
\begin{equation}
  {}^{\mathcal{W}_0}\mathbf{T}^{\text{corr}}_{\mathcal{E}_k}(t)
  =
  \Big(\Delta \mathbf{T}_{\text{pelvis}}(t)\Big)^{-1}\;
  {}^{\mathcal{W}_0}\mathbf{T}^{\text{des}}_{\mathcal{E}_k}(t),
  \label{eq:target_correction}
\end{equation}
and we pass ${}^{\mathcal{W}_0}\mathbf{T}^{\text{corr}}_{\mathcal{E}_k}(t)$ to the IK solver.
This can be interpreted either as correcting task targets or as updating an internal IK tracking frame; both are equivalent up to where the correction is applied.

\paragraph{Filtering on $SE(3)$.}
To reduce jitter from encoder noise and small contact deformation, we apply a low-pass filter on $SE(3)$ (in the Lie algebra) before compensation:
\begin{align}
\overline{\Delta \mathbf{T}}_{\text{pelvis}}(t)
&= \mathrm{LPF}\!\left(\Delta \mathbf{T}_{\text{pelvis}}(t)\right), \label{eq:lpf_delta}\\
{}^{\mathcal{W}_0}\mathbf{T}^{\text{corr}}_{\mathcal{E}_k}(t)
&= \left(\overline{\Delta \mathbf{T}}_{\text{pelvis}}(t)\right)^{-1}
\,{}^{\mathcal{W}_0}\mathbf{T}^{\text{des}}_{\mathcal{E}_k}(t).
\label{eq:lpf_comp}
\end{align}

\section{Synthetic Data Generation: Kinodynamically Validated Trajectories}
\label{app:datagen}
\label{sec:synth_data}

Our goal is to generate large-scale grasp demonstrations that are not only \emph{kinematically} valid (reachable and collision-free), but also \emph{dynamically executable} under the same hierarchical whole-body control stack used at deployment. This controller-in-the-loop filtering is key to learning kinodynamic-aware affordances for floating-base humanoids, where many geometrically plausible grasps fail due to stability and tracking limitations.

\subsection{Domain Randomization}
\label{sec:synth_dr}

\paragraph{Object assets.}
We draw object meshes from Objaverse~\citep{deitke2022objaverseuniverseannotated3d} and YCB~\citep{calli2015ycb}; OOD evaluation objects are held out from the training set. To promote shape- and scale-level generalization, each object is isotropically re-scaled by a random factor within a bounded interval.

\paragraph{Scene parameterization.}
Each scene contains a single axis-aligned box table and one target object:
\begin{equation}
\mathcal{S}=\{(\bT_{\mathrm{table}},\bs_{\mathrm{table}}),(\bT_{\mathrm{obj}},m_{\mathrm{obj}})\}.
\end{equation}
Table dimensions are sampled uniformly within predefined bounds. The table-top height is randomized relative to the workspace origin to induce diverse reach regimes:
\begin{equation}
z_{\mathrm{table}} \sim \mathcal{U}(z_{\min}, z_{\max}) + z_{\mathrm{ws}}.
\end{equation}
Tabletop heights span 0.40--0.90\,m, the training range referred to in Sec.~\ref{sec:experiments}.
The object is placed uniformly over the table surface with an edge margin. Its vertical position is set such that the lowest point of the scaled mesh rests on the tabletop, its mass is sampled as $m_{\mathrm{obj}}\sim\mathcal{U}(0.1,0.5)$\,kg, and its yaw is randomized:
\begin{equation}
\psi \sim \mathcal{U}(\psi_{\min}, \psi_{\max}).
\end{equation}

\paragraph{World-frame perturbations.}
To avoid a single canonical base-to-object transform, we additionally perturb the table center in the world frame using a run-specific random seed, and apply the corresponding rigid transform to the object pose. This produces distinct scene files across parallel collection runs while preserving valid on-table placement.

\subsection{Grasp Optimization and Motion Planning}
\label{sec:synth_opt_plan}

Given a randomized scene $\mathcal{S}$, we generate grasp demonstrations in batches.
For each scene, we sample up to $N_{\mathrm{cand}}$ grasp candidates and retain only those that are (i) kinematically feasible (reachable and collision-free) and (ii) dynamically executable under simulation evaluation (Sec.~\ref{sec:synth_dyn_filter}).

\subsubsection{Energy-Based Grasp Optimization}
\label{sec:synth_grasp_opt}

We optimize grasp configurations in joint space. Let $\bq \in \mathbb{R}^{D}$ denote the robot configuration, comprising the pelvis pose, waist, arm, and hand joints, so that the hand and the posture carrying it are optimized jointly, and let $\{\bp_i^c(\bq)\}_{i=1}^{N_c}$ be the fingertip contact points computed by forward kinematics.
We minimize an energy function
\begin{equation}
\begin{aligned}
E(\bq)
&= w_{\mathrm{dis}}E_{\mathrm{dis}}(\bq)
 + w_{\mathrm{pen}}E_{\mathrm{pen}}(\bq) \\
&\quad + w_{\mathrm{spen}}E_{\mathrm{spen}}(\bq)
 + w_{\mathrm{jnt}}E_{\mathrm{jnt}}(\bq).
\end{aligned}
\label{eq:synth_energy_main}
\end{equation}

\paragraph{Contact distance.}
We encourage designated fingertip points to lie near the object surface via the object signed distance field (SDF):
\begin{equation}
E_{\mathrm{dis}}(\bq)=\sum_{i=1}^{N_c}\left|\,\SDF_{\mathrm{obj}}\!\left(\bp_i^c(\bq)\right)\right|.
\end{equation}

\paragraph{Scene collision.}
We penalize penetration of robot collision spheres into the scene SDF:
\begin{equation}
E_{\mathrm{pen}}(\bq)=\sum_{j=1}^{N_s}\max\!\left(0,\; d_{\mathrm{pad}}-\SDF_{\mathrm{scene}}(\bs_j(\bq))\right),
\end{equation}
where $\bs_j(\bq)$ is the center of collision sphere $j$ and $d_{\mathrm{pad}}$ is a safety padding.

\paragraph{Self-collision.}
To avoid inter-finger collisions, we penalize close pairs:
\begin{equation}
E_{\mathrm{spen}}(\bq)=\sum_{(j,k)\in\mathcal{P}}\max\!\left(0,\; d_{\mathrm{self}}-\|\bs_j(\bq)-\bs_k(\bq)\|\right)^2.
\end{equation}

\paragraph{Joint limits.}
We apply a soft penalty for joint limit violation:
\begin{equation}
E_{\mathrm{jnt}}(\bq)=\sum_{i=1}^{D}\Big[\max(0,q_i^{\min}-q_i)^2+\max(0,q_i-q_i^{\max})^2\Big].
\end{equation}

\paragraph{Reachability and balance.}
The joint limits include the pelvis range the locomotion controller can realize, so every optimized posture can be commanded.
Balance is not an energy term: a static criterion does not predict the closed-loop RL policy under a reaching load.
Dynamic feasibility is verified downstream (Sec.~\ref{sec:synth_dyn_filter}).

\paragraph{Optimization.}
We solve Eq.~\eqref{eq:synth_energy_main} with a simulated-annealing-style optimizer that combines autodiff gradients with stochastic acceptance~\citep{ye2026robokit}.
At iteration $t$, a candidate $\bq'$ is proposed by an adaptive step,
\begin{equation}
\bq'=\bq-\eta(t)\cdot\frac{\nabla_{\bq}E(\bq)}{\sqrt{\bv_t}+\epsilon},
\qquad
\bv_t=\mu(\nabla_{\bq}E)^2+(1-\mu)\bv_{t-1},
\end{equation}
and accepted with probability
\begin{equation}
P(\text{accept})=\min\!\left(1,\exp\!\left(\frac{E(\bq)-E(\bq')}{T(t)}\right)\right).
\end{equation}
The step size $\eta(t)$ and temperature $T(t)$ follow exponential decay schedules (Tab.~\ref{tab:hyp_data_gen}).

\subsubsection{Motion Planning and Kinematic Filtering}
\label{sec:synth_mp}

For each optimized grasp configuration $\bq_{\mathrm{grasp}}$, we construct a multi-phase joint-space trajectory
$\{\bq(t)\}_{t=0}^{T_{\mathrm{tot}}}$ comprising: (i) retreat to a pre-grasp, (ii) approach to grasp, (iii) hold, (iv) post-grasp closure, and (v) lift.
The pre-grasp and post-grasp configurations are generated by SDF-gradient updates that move the arm (retreat) or hand joints (closure) along $\nabla \SDF_{\mathrm{scene}}$.
The lift keyframe is obtained by raising the end-effector by $\delta_z$ and solving IK; a keyframe is \emph{reachable} if this IK converges within joint limits, so kinematic validity is a property of the whole approach, grasp, and lift motion rather than of a static grasp pose.
Intermediate segments are initialized by interpolation and refined by trajectory smoothing.

\paragraph{Trajectory smoothing.}
We optimize the full trajectory by minimizing a collision penalty and a smoothness term:
\begin{equation}
\mathcal{L}_{\mathrm{traj}}=
\mathcal{L}_{\mathrm{pen}}+
\lambda_{\mathrm{smooth}}\mathcal{L}_{\mathrm{smooth}},
\end{equation}
where
\begin{equation}
\mathcal{L}_{\mathrm{pen}}=
\frac{1}{T_{\mathrm{tot}}N_s}\sum_{t,j}
\left[\max\!\left(0,\; d_{\mathrm{pad}}-\SDF_{\mathrm{scene}}(\bs_j(t))\right)\right]^2,
\end{equation}
and
\begin{equation}
\mathcal{L}_{\mathrm{smooth}}=\sum_{i=1}^{D}w_i\Big(\mathrm{Std}(\Delta q_i)+\alpha_{\mathrm{acc}}\mathrm{Std}(\Delta^2 q_i)\Big).
\end{equation}

\paragraph{Kinematic feasibility.}
We treat a sample as \emph{kinematically feasible} only if it passes both the grasp optimization stage and the motion planning stage.
For a configuration $\bq$, define the scene penetration of collision sphere $j$ as
\begin{equation}
\mathrm{pen}_j(\bq)\triangleq\max\!\left(0,\; d_{\mathrm{pad}}-\SDF_{\mathrm{scene}}(\bs_j(\bq))\right),
\end{equation}
and the fingertip-to-object surface distance for contact point $i$ as
\begin{equation}
\mathrm{dis}_i(\bq)\triangleq\left|\SDF_{\mathrm{obj}}\!\left(\bp_i^c(\bq)\right)\right|.
\end{equation}
Grasp optimization succeeds if every keyframe is reachable and the optimized grasp configuration $\bq_{\mathrm{grasp}}$ satisfies
\begin{equation}
\texttt{opt\_success}
=\mathbb{1}\!\left[
\begin{aligned}
&\max_{j}\ \mathrm{pen}_{j}(\bq_{\mathrm{grasp}})\le \epsilon_{\mathrm{col}}
\\
&\wedge\ \max_{i}\ \mathrm{dis}_{i}(\bq_{\mathrm{grasp}})\le \epsilon_{\mathrm{dis}}
\end{aligned}
\right].
\end{equation}

Motion planning succeeds if the planned trajectory $\{\bq(t)\}_{t=0}^{T_{\mathrm{tot}}}$ remains collision-free (with padding) along the entire execution:
\begin{equation}
\texttt{mp\_success}=\mathbb{1}\!\left[\max_{t,j}\mathrm{pen}_j(\bq(t))\le \epsilon_{\mathrm{traj}}\right].
\end{equation}
A sample is kinematically valid iff
\begin{equation}
\texttt{kin\_success}=\texttt{opt\_success}\ \wedge\ \texttt{mp\_success}.
\end{equation}
Only trajectories with $\texttt{kin\_success}=1$ are passed to the dynamics-aware feasibility filtering stage.

Optimization and planning schedules are listed in Tab.~\ref{tab:hyp_data_gen}.

\subsection{Dynamics-Aware Feasibility Filtering}
\label{sec:synth_dyn_filter}

\paragraph{Why dynamics filtering?}
For floating-base humanoids, kinematic feasibility does not guarantee executability: trajectories can violate stability constraints, exceed controller tracking capability, or induce unstable whole-body transients. We therefore apply a controller-in-the-loop dynamics check to retain only trajectories that are feasible under the deployment control stack.

\paragraph{Massively parallel controller replay.}
We replay each kinematically feasible trajectory in ManiSkill3~\citep{tao2025maniskill3gpuparallelizedrobotics}, a GPU-parallel physics simulator built on SAPIEN~\citep{xiang2020sapien}, at the deployment control rate (50\,Hz). Crucially, the same hierarchical whole-body control structure as real deployment is used during replay (whole-body IK producing upper-body targets and low-level locomotion control executing the commanded motion).

\paragraph{Dynamic feasibility criteria.}
A trajectory is marked \texttt{sim\_success} if it satisfies:
\begin{enumerate}[nosep,leftmargin=1.2em]
\item \textbf{Task success:} the object is lifted by at least $\delta_{\mathrm{lift}}=0.05\,\mathrm{m}$ and remains in hand.
\item \textbf{Stable execution:} no fall/termination, and base attitude remains bounded, $|\mathrm{roll}|,|\mathrm{pitch}|\le \theta_{\max}$.
\item \textbf{Trackability:} tracking error of commanded end-effector keyframes stays below a threshold, $\max_t \| \bp_{\mathrm{ee}}(t)-\bp^*_{\mathrm{ee}}(t)\|\le e_{\max}$, with no abrupt spikes in commanded motion.
\end{enumerate}
We treat $(\theta_{\max}, e_{\max})$ as fixed thresholds chosen to match the tolerances of the deployment controller.

\paragraph{Final acceptance mask.}
We retain a sample only if it passes \emph{both} kinematic and dynamics feasibility:
\begin{equation}
\texttt{success}[i]=\texttt{kin\_success}[i]\ \wedge\ \texttt{sim\_success}[i].
\label{eq:synth_mask}
\end{equation}
By construction, the resulting dataset is supported on actions that are executable under whole-body kinodynamic constraints, enabling the learned policy to internalize kinodynamic-aware affordances rather than purely geometric ones.
Fig.~\ref{tab:filtering} profiles an example collection run (508M candidates to 18K accepted grasps); its survival rates are representative of every run, and the full training set aggregates all parallel runs with distinct seeds into 600k trajectories.

\subsection{Rendering and View Selection}
\label{sec:synth_render}

We render object-centric point cloud observations for each accepted sample using a GPU-parallel renderer (ManiSkill3~\citep{tao2025maniskill3gpuparallelizedrobotics}). For each scene, we render $V=16$ camera views with small RPY perturbations around a calibrated nominal extrinsic, and extract an \emph{object-only} segmented point cloud of $N_{\mathrm{pts}}=4096$ points, randomly downsampled to $N=2048$ for the policy.

\paragraph{View selection.}
Among candidate views whose frustums contain the object 3D bounding box, we select the view maximizing the number of valid (nonzero) 3D points:
\begin{equation}
v^*=\arg\max_{v\in\mathcal{V}_{\mathrm{visible}}}\sum_{i=1}^{N_{\mathrm{pts}}}\mathbb{1}\!\left[\|\bp_i^{(v)}\|>\epsilon\right].
\end{equation}
If no view passes the visibility check, the scene is discarded to avoid degenerate observations. At deployment, this offline oracle view selection is replaced by the online active perception module in the main paper.

\section{Policy Architecture}
\label{app:policy}
\label{sec:policy}

\paragraph{Overview.}
We learn a conditional generative grasp policy that maps a segmented object point cloud
$\mathcal{P}\in\mathbb{R}^{N\times 3}$ to a \emph{distribution} over multi-stage end-effector (EE) actions.
Let $\mathbf{f}=\mathrm{PointNet}(\mathcal{P})\in\mathbb{R}^{d_f}$ be a global shape feature.
Our policy is a conditional VAE (CVAE) with latent $\bz\in\mathbb{R}^{D_z}$, defining
\begin{equation}
p_\theta(\ba \mid \mathcal{P})
=
\int p_\theta(\ba \mid \bz,\mathbf{f})\,p(\bz)\,d\bz,
\qquad p(\bz)=\mathcal{N}(\mathbf{0},\mathbf{I}),
\end{equation}
where $\ba\in\mathbb{R}^{T\times D_a}$ is a $T$-stage grasp action (pre-grasp, grasp, post-grasp).
All architectural and training hyperparameters are summarized in Sec.~\ref{sec:hyperparams}.

\paragraph{Action parameterization.}
Each stage action $\ba_t\in\mathbb{R}^{D_a}$ consists of EE translation, EE orientation as a unit quaternion,
and hand joint targets:
\begin{equation}
\ba_t \triangleq [\,\bp_t;\,\br_t;\,\bh_t\,],\qquad
\ba \triangleq [\ba_1;\ldots;\ba_T]\in\mathbb{R}^{T\times D_a}.
\end{equation}

\subsection{Vision Encoder: PointNet}
\label{sec:pointnet}

We encode $\mathcal{P}$ using a PointNet-style~\citep{qi2017pointnet} encoder:
\begin{equation}
\mathbf{f} = \phi\!\Big(\max\nolimits_{n}\, h(\mathcal{P})\Big),
\end{equation}
where $h(\cdot)$ is a shared-point MLP (implemented by stacked $\mathrm{Conv1D}{+}\mathrm{ReLU}$ blocks),
$\max_n$ is channel-wise max-pooling over points, and $\phi(\cdot)$ projects to dimension $d_f$.
We omit BatchNorm in the encoder to avoid train/test mismatch under small-batch deployment.

\subsection{Conditional VAE (CVAE)}
\label{sec:cvae}

Let $\ba^{\mathrm{flat}}\in\mathbb{R}^{T\cdot D_a}$ be the flattened action.
During training, the encoder models an approximate posterior
$q_\varphi(\bz \mid \ba^{\mathrm{flat}}, \mathbf{f})=\mathcal{N}(\bmu,\mathrm{diag}(\bsigma^2))$:
\begin{equation}
[\bmu,\ \log\bsigma^2]
=
\mathrm{MLP}_{\mathrm{enc}}\big([\ba^{\mathrm{flat}};\mathbf{f}]\big),
\qquad
\bz = \bmu + \bsigma \odot \bepsilon,\ \ \bepsilon\sim\mathcal{N}(\mathbf{0},\mathbf{I}),
\end{equation}
where $\log\bsigma^2$ is clamped to a fixed range (Sec.~\ref{sec:hyperparams}) for numerical stability.
The decoder predicts the action:
\begin{equation}
\hat{\ba}^{\mathrm{flat}} = \mathrm{MLP}_{\mathrm{dec}}\big([\bz;\mathbf{f}]\big),
\qquad
\hat{\ba}=\mathrm{reshape}(\hat{\ba}^{\mathrm{flat}})\in\mathbb{R}^{T\times D_a}.
\end{equation}

\subsection{Training Objective and Inference}
\label{sec:training_loss}

We minimize the negative evidence lower bound (ELBO):
\begin{equation}
\mathcal{L}
=
\underbrace{\mathbb{E}_{q_\varphi(\bz\mid \ba^{\mathrm{flat}},\mathbf{f})}\!\left[\mathcal{L}_{\mathrm{rec}}(\hat{\ba},\ba)\right]}_{\text{reconstruction}}
+
w_{\mathrm{KL}}
\underbrace{D_{\mathrm{KL}}\!\left(q_\varphi(\bz\mid \ba^{\mathrm{flat}},\mathbf{f})\ \|\ \mathcal{N}(\mathbf{0},\mathbf{I})\right)}_{\text{regularization}}.
\end{equation}
We use an $\ell_1$ reconstruction loss over translation, quaternion, and hand joints across stages.
Quaternions are normalized and sign-canonicalized to resolve $\br\equiv-\br$:
\begin{equation}
\bar{\br}\leftarrow \br/\|\br\|,\quad
\hat{\bar{\br}}\leftarrow \hat{\br}/\|\hat{\br}\|,\quad
\hat{\bar{\br}}\leftarrow \mathrm{sign}(\langle\hat{\bar{\br}},\bar{\br}\rangle)\cdot \hat{\bar{\br}}.
\end{equation}

\paragraph{Inference.}
At test time, we discard the encoder and draw $K$ latents $\bz\sim\mathcal{N}(\mathbf{0},\mathbf{I})$ for the alignment stage (Sec.~\ref{sec:alignment}),
decoding each as $\hat{\ba}=\mathrm{MLP}_{\mathrm{dec}}([\bz;\mathrm{PointNet}(\mathcal{P})])$.
The predicted EE targets are executed by the downstream whole-body stack (IK, collision-aware planning, and controller) as in the main paper.

\section{Hyperparameter Summary}
\label{app:hyperparams}
\label{sec:hyperparams}

\begin{table}[H]
\centering
\small
\setlength{\tabcolsep}{6pt}
\renewcommand{\arraystretch}{1.05}
\caption{Key hyperparameters for synthetic grasp generation (optimization and motion planning).}
\label{tab:hyp_data_gen}
\begin{tabular}{@{}llc l@{}}
\toprule
\textbf{Stage} & \textbf{Parameter} & \textbf{Symbol} & \textbf{Value} \\
\midrule
\multicolumn{4}{@{}l@{}}{\textbf{Grasp Optimization (Simulated Annealing)}} \\
Iterations & Number of iterations & $N_{\mathrm{iter}}$ & 5{,}000 \\
Batching & Grasp samples per scene & $N_{\mathrm{cand}}$ & 8{,}192 \\
Contacts & Fingertip contact points & $N_c$ & 4 \\
Energy & Contact-distance weight & $w_{\mathrm{dis}}$ & 100 \\
Energy & Scene-penetration weight & $w_{\mathrm{pen}}$ & 100 \\
Energy & Self-penetration weight & $w_{\mathrm{spen}}$ & 50 \\
Energy & Joint-limit weight & $w_{\mathrm{jnt}}$ & 1 \\
Steps & Initial step size & $\eta_0$ & 0.005 \\
Steps & Step decay period & $\tau_\eta$ & 50 \\
Temp. & Initial temperature & $T_0$ & 18 \\
Temp. & Temperature decay rate & $\gamma_T$ & 0.95 \\
Temp. & Annealing period & $\tau_T$ & 30 \\
Init. & Front approach angle range & $\theta_{\mathrm{front}}$ & $[-\pi/2,\;-\pi/10]$ \\
Init. & Approach distance range & $d$ & $[0.02,\;0.15]$\,m \\
Init. & Hand template (6-DoF) & $\bmu_{\mathrm{hand}}$ & $(0.8,\;0.05,\;0.6,\;0.6,\;0.6,\;0.6)$ \\
\midrule
\multicolumn{4}{@{}l@{}}{\textbf{Motion Planning (Pre/Grasp/Post + Smoothing)}} \\
Phases & Init$\rightarrow$Pre-grasp steps & $T_{\mathrm{init}}$ & 150 \\
Phases & Pre-grasp$\rightarrow$Grasp steps & $T_{\mathrm{pg}}$ & 80 \\
Phases & Holding steps & $T_{\mathrm{hold}}$ & 15 \\
Phases & Grasp$\rightarrow$Post-grasp steps & $T_{\mathrm{gp}}$ & 50 \\
Phases & Lift height & $\delta_z$ & 0.05\,m \\
Collision & Safety padding & $d_{\mathrm{pad}}$ & 0.025\,m \\
Heuristic & Retreat step factor (arm) & $w_{\mathrm{arm}}$ & 0.001 \\
Heuristic & Closure factor (hand) & $w_{\mathrm{hand}}$ & 0.8 \\
Smoothing & Arm smoothness weight & $w_{\mathrm{sm}}^{\mathrm{arm}}$ & 100 \\
Smoothing & Hand smoothness weight & $w_{\mathrm{sm}}^{\mathrm{hand}}$ & 10 \\
Batching & Trajectories per batch & $B_{\mathrm{traj}}$ & 1{,}024 \\
\bottomrule
\end{tabular}
\end{table}

\begin{table}[H]
\centering
\small
\setlength{\tabcolsep}{6pt}
\renewcommand{\arraystretch}{1.05}
\caption{Hyperparameters for policy learning, evaluation, and rendering.}
\label{tab:hyp_policy_eval}
\begin{tabular}{@{}llc l@{}}
\toprule
\textbf{Module} & \textbf{Parameter} & \textbf{Symbol} & \textbf{Value} \\
\midrule
\multicolumn{4}{@{}l@{}}{\textbf{Policy (PointNet + CVAE)}} \\
Input & Policy input point cloud size & $N$ & 2{,}048 \\
Action & Stages (pre/grasp/post) & $T$ & 3 \\
Action & Action dimension & $D_a$ & 13 \\
Action & Hand DoF in action & $D_h$ & 6 \\
Encoder & PointNet feature dimension & $d_f$ & 512 \\
Latent & CVAE latent dimension & $D_z$ & 256 \\
MLP & Encoder/decoder hidden dims & --- & $[256,\;256,\;256]$ \\
MLP & Activation / normalization & --- & SiLU / LayerNorm \\
Stability & $\log\bsigma^2$ clamp range & $[\ell_{\min},\ell_{\max}]$ & $[-10,\;10]$ \\
Loss & KL weight & $w_{\mathrm{KL}}$ & 0.1 \\
Optim. & Optimizer & --- & AdamW \\
Optim. & Learning rate & $\eta_{\mathrm{lr}}$ & $10^{-4}$ \\
Optim. & Weight decay & $\lambda_{\mathrm{wd}}$ & $10^{-4}$ \\
Batch & Batch size (scenes) & $B$ & 128 \\
Sampling & Actions per scene & $N_a$ & 12 \\
Schedule & LR decay step (iter) & $t_{\mathrm{decay}}$ & 66{,}667 \\
Schedule & LR decay factor & $\gamma_{\mathrm{lr}}$ & 0.1 \\
Training & Total iterations & $T_{\mathrm{iter}}$ & 100{,}000 \\
Split & Train/test split (scenes) & --- & 97\% / 3\% \\
Eval. & Evaluation frequency & --- & every 10{,}000 iters \\
\midrule
\multicolumn{4}{@{}l@{}}{\textbf{Simulation Evaluation}} \\
Engine & Physics simulator & --- & ManiSkill3 (SAPIEN) \\
Parallelism & Parallel environments & $N_{\mathrm{env}}$ & 512 \\
Control & Control frequency & $f_{\mathrm{ctrl}}$ & 50\,Hz \\
Object & Object mass & $m_{\mathrm{obj}}$ & $\mathcal{U}(0.1,\,0.5)$\,kg \\
Metric & Lift threshold & $\delta_{\mathrm{lift}}$ & 0.05\,m \\
Contact & Friction (dyn/static) & --- & 2.0 / 2.0 \\
Contact & Restitution & --- & 0.0 \\
\midrule
\multicolumn{4}{@{}l@{}}{\textbf{Rendering / View Selection}} \\
Views & Views per scene & $V$ & 16 \\
Points & Rendered point cloud size (before downsample) & $N_{\mathrm{raw}}$ & 4{,}096 \\
Points & Policy input points (after downsample) & $N$ & 2{,}048 \\
Jitter & Roll jitter & $\Delta\phi$ & $\pm 5^\circ$ \\
Jitter & Pitch jitter & $\Delta\theta$ & $\pm 8^\circ$ \\
Jitter & Yaw jitter & $\Delta\psi$ & $\pm 12^\circ$ \\
Seg. & Segment extraction & --- & object-only \\
\bottomrule
\end{tabular}
\end{table}

\section{Extended Experimental Analysis}
\label{app:extended_exp}
\label{sec:supp_extended_exp}

This section collects supplementary figures and analyses referenced
from Sec.~\ref{sec:experiments} of the main paper: real-world qualitative
rollouts, visual failure modes, scene generalization, the training-data
scaling curve, and the full results of the large-scale simulation
evaluation (spatial success-rate fields and per-object success rates).

\subsection{Real-World Qualitative Rollouts}
\label{sec:supp_qualitative}

\begin{figure}[h]
    \centering
    \includegraphics[width=\linewidth]{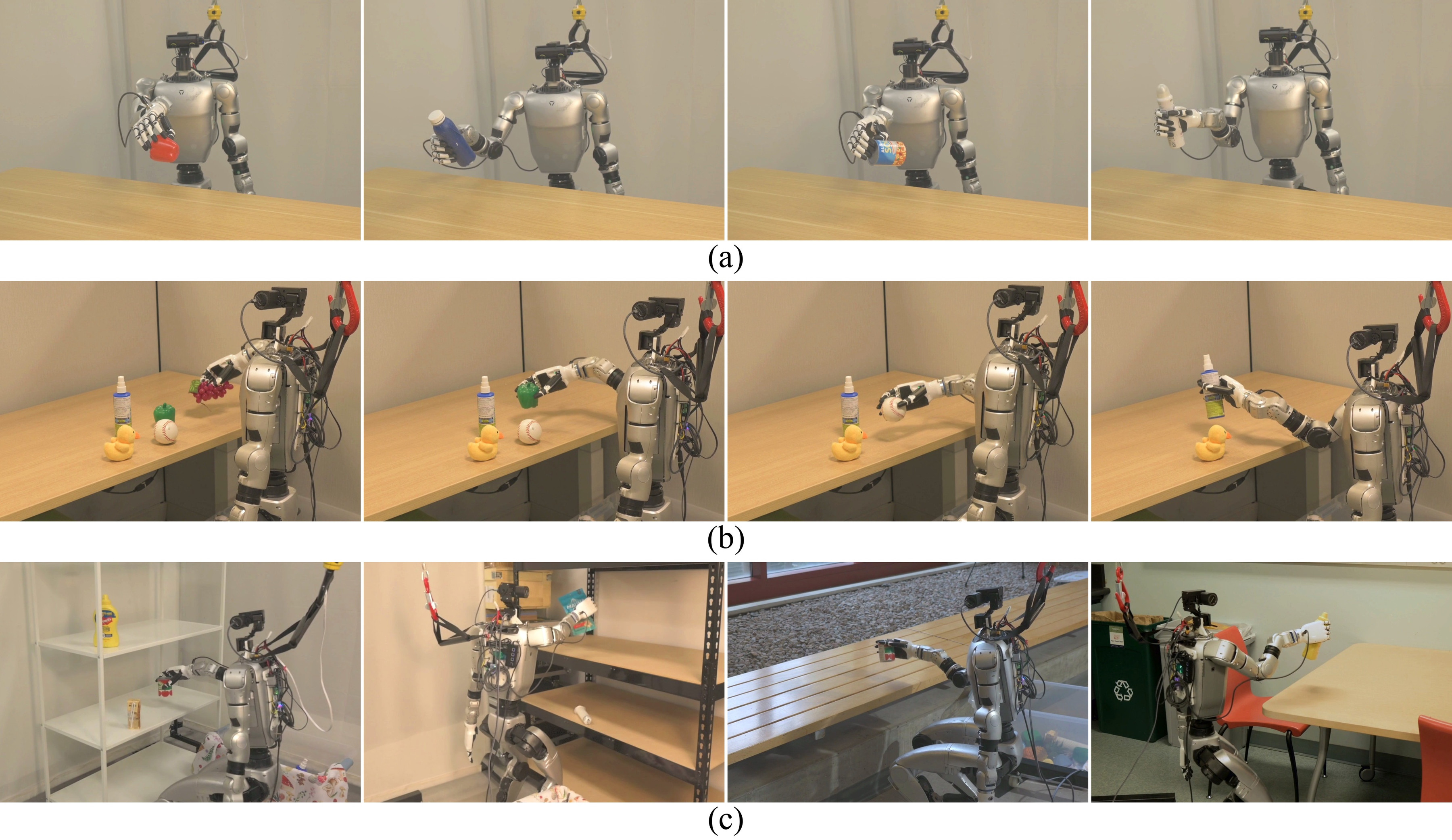}
    \caption{Real-world rollouts, zero-shot from simulation.
    (a)~Tabletop grasping of diverse objects;
    (b)~long-reach grasping across a table;
    (c)~deployment in varied unseen environments and heights (shelves, benches, tables).}
    \label{fig:real-demo}
\end{figure}

Fig.~\ref{fig:real-demo} shows representative successes across diverse objects, reaches, and scenes: tabletop manipulation~(a), long-reach grasping~(b), and zero-shot deployment in unseen environments at varied heights~(c). Across all settings the policy coordinates active perception, whole-body motion, and dexterous manipulation with no scene-specific tuning.

\subsection{Visual Failure Modes}
\label{sec:supp_failure_viz}

\begin{figure}[!t]
    \centering
    \includegraphics[width=\columnwidth]{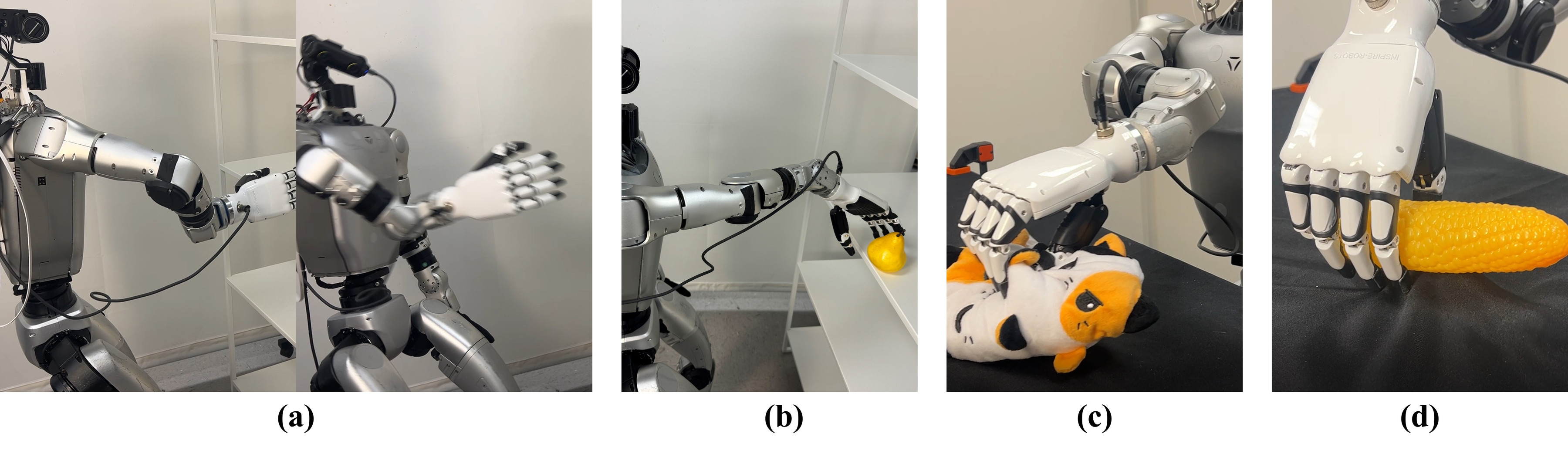}
    \caption{Failure modes eliminated by our pipeline.
    (a)~No dynamics filter: unstable jittering.
    (b)~No WB co-optimization/alignment: twisted infeasible poses.
    (c)~GraspNet: unnatural gripper-designed grasps.
    (d)~No WB--hand co-optimization: hand--environment collisions.}
    \label{fig:supp_failure}
\end{figure}

Fig.~\ref{fig:supp_failure} visualizes the four representative failure
modes that the Full pipeline eliminates relative to its ablations and
to the gripper-based baseline. (a) and (b) are the kinematically valid
but dynamically infeasible configurations that the dynamics filter
rejects; (c) and (d) are the morphology-mismatched and collision-prone
grasps produced when whole-body and hand DoFs are not jointly
optimized.

\subsection{Scene Generalization across Real-World Environments}
\label{sec:supp_scene_gen}

\begin{figure}[!htbp]
    \centering
    \includegraphics[width=1.0\linewidth]{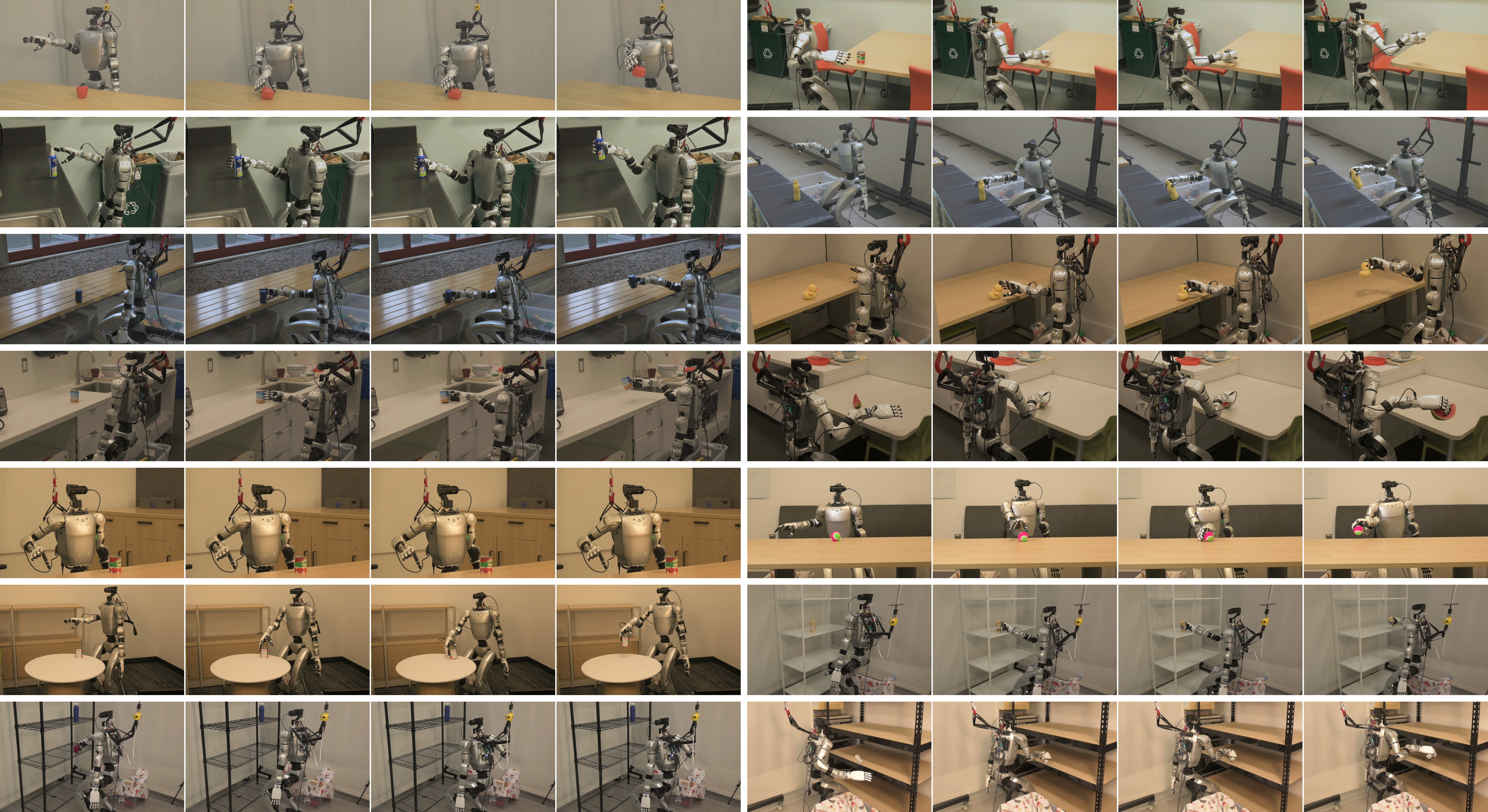}
    \caption{Our full pipeline in fourteen real-world environments.}
    \label{fig:supp_scene_gen}
\end{figure}

Beyond the per-height generalization reported in the main Analysis (Sec.~\ref{sec:spatial}), Fig.~\ref{fig:supp_scene_gen} shows the same policy operating across different real-world environments: kitchens, hallways with railings, multi-tier shelves, dining tables, and open lab benches at varying heights. The policy is trained purely on randomized tabletop scenes in simulation and uses only the egocentric RGB-D and active perception stack at deployment; no scene-specific fine-tuning, per-scene demonstrations, or environment annotation is required. Robustness to surface geometry and clutter follows from the diversity of the training distribution rather than any scene-aware module; lighting and background variation are absorbed by the open-vocabulary detection and segmentation front end, since the policy consumes only the segmented object point cloud.

\subsection{Synthetic Data Scaling}
\label{sec:supp_scaling}

\begin{figure}[!t]
    \centering
    \includegraphics[width=\columnwidth]{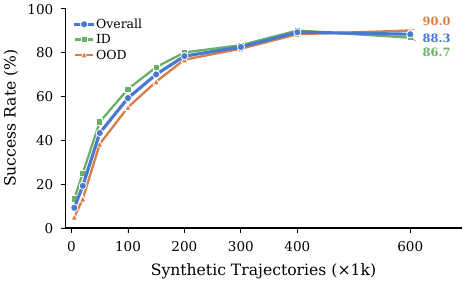}
    \caption{Data scaling (real world, oracle perception, 5\,ID + 5\,OOD
    objects; 120 trials per data scale, a separate protocol from Task~2).
    SR scales from 9.2\% at 5k trajectories to ${\sim}$89\% at 400k and
    plateaus thereafter, with OOD benefiting most.}
    \label{fig:supp_scaling}
\end{figure}

Fig.~\ref{fig:supp_scaling} reports real-world SR as a function of
the synthetic training-data volume. Each data scale is evaluated with
oracle perception on 120 trials (60 ID, 60 OOD), for 1{,}080 real-world trials over nine data scales; this protocol is separate from Task~2's 150-trial
height sweep, so absolute values are not directly comparable to
Fig.~\ref{fig:spatial}. SR rises monotonically from
9.2\% at 5k trajectories to 89.2\% at 400k and plateaus thereafter.
OOD objects benefit most (5.0\%$\to$90.0\%) while ID saturates
earlier; the relative OOD degradation narrows from 62\% at 5k to
under 2\% above 300k, indicating that data diversity progressively
closes the generalization gap to novel objects.

\paragraph{Computational cost.}
Offline generation: ${\sim}$700 GPU-hours for 600k trajectories
(fully parallelizable); CVAE training: ${\sim}$1\,h on a single GPU.
At deployment (RTX~5090): perception (DINO\,+\,SAM) 148\,ms, CVAE
inference 0.4\,ms, motion planning ${\sim}$200\,ms, WB-IK ${\sim}$1\,ms per batched solve over all $K$ candidates.
End-to-end grasp attempts take 15--40\,s, dominated by physical
robot motion.

\subsection{Large-Scale Simulation Evaluation}
\label{sec:supp_sim_eval}

This section reports the full results of the Task~3 simulation evaluation
(protocol in Sec.~\ref{sec:exp_setup}; band aggregates and SR-vs-height
curves in Fig.~\ref{fig:sim_eval}): the spatial success-rate fields (side
and top view) and per-object success rates.

\paragraph{Protocol details.}
The evaluation runs in ManiSkill3~\citep{tao2025maniskill3gpuparallelizedrobotics} with the deployment whole-body controller in the loop.
Each condition sweeps 17 table heights (0.10--0.90\,m in 5\,cm steps) with
randomized planar object placement ($x\in[0.25,0.55]$\,m,
$y\in[-0.175,0.125]$\,m) and randomized yaw, over the YCB (12 objects) and
RoboCasa~\citep{nasiriany2024robocasa} (11 objects) catalogs.
Scene sets are \emph{identical} across the six variants, so all deltas are
paired. Training tables span 0.40--0.90\,m; heights 0.10--0.35\,m evaluate
extrapolation below the training range.

\paragraph{Success criterion.}
A trial succeeds only if it is \emph{held and upright}: the object is lifted
$\ge 0.05$\,m and remains lifted after a 2\,s hold with $\le 0.03$\,m of
in-hand slip, while the robot stays upright (pelvis height $\ge 0.20$\,m,
base tilt $\le 30^\circ$, and foot drift $\le 0.05$\,m). Scoring is a pure
function of the recorded telemetry, with no manual judgment. This criterion is
stricter than the real-world one (lift $\ge$5\,cm and stably held), since it
additionally penalizes post-grasp slip and any loss of balance during the
hold; simulation success rates are therefore conservative relative to the
hardware tasks.

\begin{figure}[!t]
    \centering
    \includegraphics[width=\linewidth]{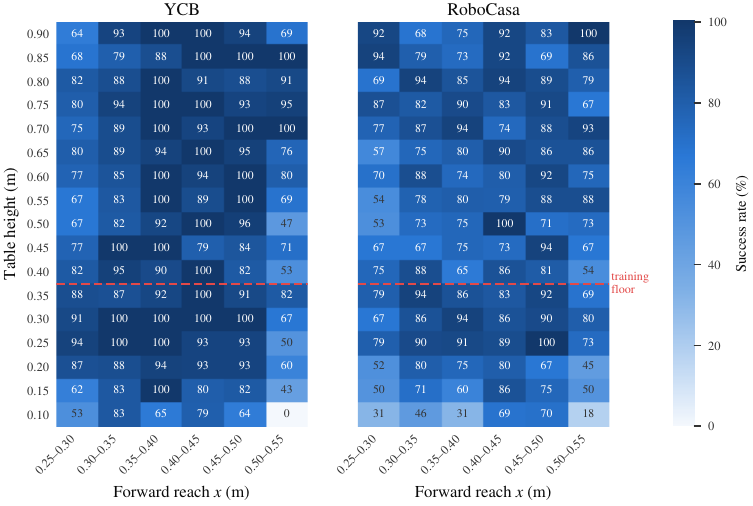}
    \caption{\textbf{Side-view success-rate field} (Full pipeline): table
    height vs.\ forward reach $x$, per catalog. The dashed line marks the
    0.40\,m training floor. Success stays high across the whole
    $0.10$--$0.90 \times 0.25$--$0.55$\,m envelope and degrades only at the
    far-reach column and the deepest squat row; per-cell $n$ varies with
    random placement (8--42), so the field should be read as a whole rather
    than cell-by-cell.}
    \label{fig:supp_sideview}
\end{figure}

\begin{figure}[!t]
    \centering
    \includegraphics[width=\linewidth]{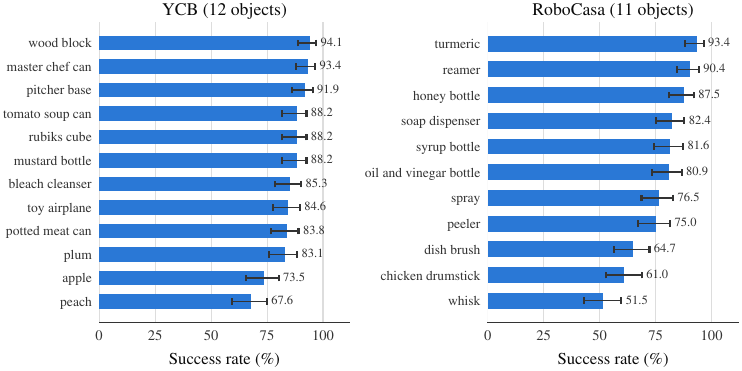}
    \caption{\textbf{Per-object success rate} (Full pipeline, $n{=}136$
    trials/object, 95\% Wilson CIs). The RoboCasa catalog's broader shape
    distribution widens the outcome spread (51.5--93.4\% vs.\ 67.7--94.1\%
    for YCB); the hardest objects are thin and elongated (whisk, chicken
    drumstick, dish brush), whose narrow graspable surfaces demand
    centimeter-precise hand placement.}
    \label{fig:supp_perobject}
\end{figure}

\paragraph{Spatial envelope and object difficulty.}
The success field over height\,$\times$\,reach (Fig.~\ref{fig:supp_sideview}) is
dense throughout the $0.10$--$0.90 \times 0.25$--$0.55$\,m envelope and thins
mainly along the far-reach column ($x\ge0.50$\,m), most severely when far reach is
\emph{combined with} deep squat, where extension
and balance compete. Across objects (Fig.~\ref{fig:supp_perobject}), the
broader RoboCasa shape distribution costs ${\sim}$8\,pts in-range
(87.4\,$\to$\,79.8\%) and widens the per-object spread (51.5--93.4\% vs.\
67.7--94.1\% for YCB); the hardest objects are thin or elongated (whisk,
chicken drumstick, dish brush) and small near-spherical fruit (peach, apple),
which demand centimeter-precise closure. The two degradations are close to
additive: YCB in-range (87.4\%) to RoboCasa extrapolation (71.4\%) costs
16\,pts, within ${\sim}$2\,pts of the sum of the single-axis costs
(6.3\,+\,7.6), so there is no catastrophic interaction.

\begin{figure}[!t]
    \centering
    \includegraphics[width=\linewidth]{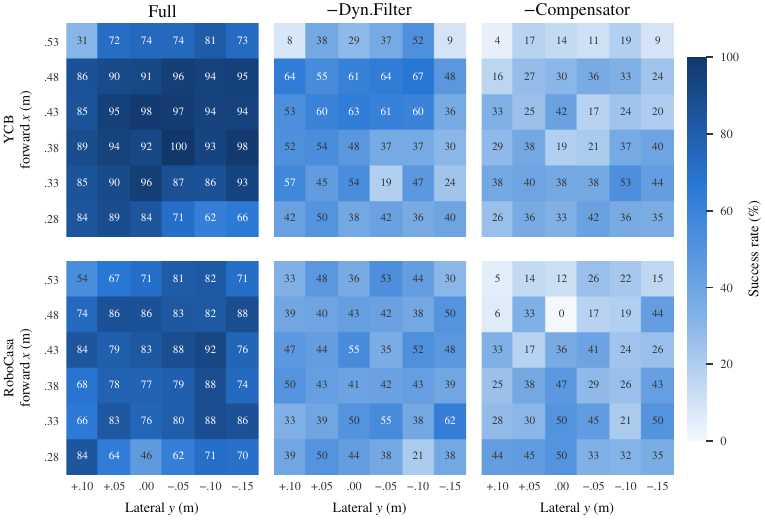}
    \caption{\textbf{Top-view success-rate fields} (plan view; robot at the
    bottom, grasping-hand side to the right): Full vs.\ --Dyn.\ Filter vs.\
    --Compensator on both catalogs, aggregated over all heights. The Full
    envelope is broad and dense; removing the dynamics filter keeps coverage
    but at uniformly lower success density; removing the compensator degrades
    the whole field, worst at far cells.}
    \label{fig:supp_topview}
\end{figure}

\end{document}